\documentclass[sigconf]{acmart}

\AtBeginDocument{%
  }

\acmDOI{XXXXXXX.XXXXXXX}
\acmConference[KDD '25]{Make sure to enter the correct
  conference title from your rights confirmation email}{Augest 03 -Augest 07, 2025}{Toronto, Canada}
\usepackage{balance}
\usepackage{colortbl}
\usepackage{subfigure}
\usepackage{listings}

\usepackage{algorithm,algcompatible}
\algnewcommand\algorithmicreturn{\textbf{return}}
\algnewcommand\RETURN{\algorithmicreturn}

\usepackage{multirow} 
\usepackage{enumitem} 
\usepackage{amsmath}
\usepackage{xcolor}
\usepackage{mwe}
\usepackage{booktabs}
\usepackage{threeparttable}
\usepackage{amsmath}
\usepackage{amsfonts}
\usepackage{flushend}

\begin{document}

\title{Diffmv: A Unified Diffusion Framework for Healthcare Predictions with Random Missing Views and View Laziness}



\author{Chuang Zhao}
\affiliation{%
  \institution{The Hong Kong University of Science and Technology }
  \city{Hong Kong}
  \country{Hong Kong}}
  \email{czhaobo@connect.ust.hk}
  
  \author{Hui Tang}
\affiliation{%
  \institution{The Hong Kong University of Science and Technology }
  \city{Hong Kong}
  \country{Hong Kong}}
  \email{eehtang@ust.hk}

  \author{Hongke Zhao}
\affiliation{%
  \institution{College of Management and Economics \& Laboratory of Computation and Analytics of Complex Management Systems (CACMS), Tianjin University;\\
  ai-deepcube.com}
  \city{Tianjin}
  \country{China}}
\email{hongke@tju.edu.cn}

  \author{Xiaomeng Li}
  \authornote{Xiaomeng Li is the corresponding author.}
\affiliation{%
  \institution{The Hong Kong University of Science and Technology }
  \city{Hong Kong}
  \country{Hong Kong}}
  \email{eexmli@ust.hk}

\renewcommand{\shortauthors}{Chuang Zhao, Hui Tang, Hongke Zhao, and Xiaomeng Li}
\begin{abstract}
Advanced healthcare predictions offer significant improvements in patient outcomes by leveraging predictive analytics. Existing works primarily utilize various views of Electronic Health Record (EHR) data, such as diagnoses, lab tests, or clinical notes, for model training. These methods typically assume the availability of complete EHR views and that the designed model could fully leverage the potential of each view. However, in practice, random missing views and view laziness present two significant challenges that hinder further improvements in multi-view utilization.
 To address these challenges, we introduce Diffmv, an innovative \underline{diff}usion-based generative framework designed to advance the exploitation of \underline{m}ultiple \underline{v}iews of EHR data. Specifically, to address random missing views, we integrate various views of EHR data into a unified diffusion-denoising framework, enriched with diverse contextual conditions to facilitate progressive alignment and view transformation. To mitigate view laziness, we propose a novel reweighting strategy that assesses the relative advantages of each view, promoting a balanced utilization of various data views within the model. Our proposed strategy achieves superior performance across multiple health prediction tasks derived from three popular datasets, including multi-view and multi-modality scenarios.

\end{abstract}

\ccsdesc[500]{Applied computing~Health informatics}

\keywords{Diffusion model, Healthcare prediction, Missing views}

%

\maketitle

\section{Introduction}\label{sec:intro}
Healthcare prediction plays a critical role in improving patient outcomes through tasks such as phenotype prediction~\cite{xu2024flexcare,zhao2025beyond} and length-of-stay estimation~\cite{jiang2023graphcare,yang2023pyhealth}. For instance, accurate phenotype prediction enables early intervention, allowing healthcare providers to tailor personalized treatment plans to individual patient characteristics~\cite{nguyen2022deep}. However, healthcare prediction is a double-edged sword; erroneous predictions can lead to misdiagnoses and inappropriate treatments, resulting in adverse health outcomes and increased healthcare costs~\cite{gao2024precision,zhang2022m3care}. This underscores the urgent need for more robust and advanced predictive tools. 

Current research primarily focuses on extracting patient representations from diverse views of Electronic Health Record (EHR) data~\cite{zhao2025unveiling,boll2024graph}, broadly categorized into three paradigms: rule-based, instance-based, and longitudinal-based methods. Rule-based methods~\cite{stone2022systematic} employ predefined rules and if-then flow charts to derive outcomes. While effective in certain contexts, they often lack adaptability to novel cases and incur high manual costs due to the need for continuous updating of rules. Instance-based methods~\cite{bhoi2021personalizing,wang2024recent}, on the other hand, concentrate on the current state of the patient by utilizing machine learning models or graph neural networks to capture EHR interactions. Although these methods are entirely data-driven, they often struggle with accuracy because they underutilize historical visit data, missing valuable longitudinal insights~\cite{zhao2024enhancing,xu2023seqcare,xu2024protomix}. Longitudinal methods incorporate patients' historical health records, providing a more holistic understanding of their conditions over time. Recently, there has been a growing interest in integrating multi-view EHR data to capture additional semantic information~\cite{kim2023clinical,wu2024multimodal}. For clarity, as in~\cite{zhu2024prism,yu2024smart,xu2024flexcare}, we consider different types of EHR data as distinct views, enhancing the model's ability to learn from diverse data sources.

\begin{figure}[!t]
  \centering
\setlength{\abovecaptionskip}{-0.05cm}   
\setlength{\belowcaptionskip}{-0.1cm}   
\subfigure[Random Missing View]{ 
\begin{minipage}[t]{0.32\linewidth}
\centering
\includegraphics[width=\linewidth,height=0.62\linewidth]{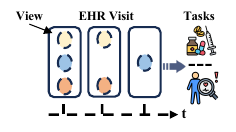}
\label{fig:motiv:missing}
\end{minipage}%
}%
\subfigure[Missing Status]{ 
\begin{minipage}[t]{0.32\linewidth}
\centering
\includegraphics[width=\linewidth,height=0.62\linewidth]{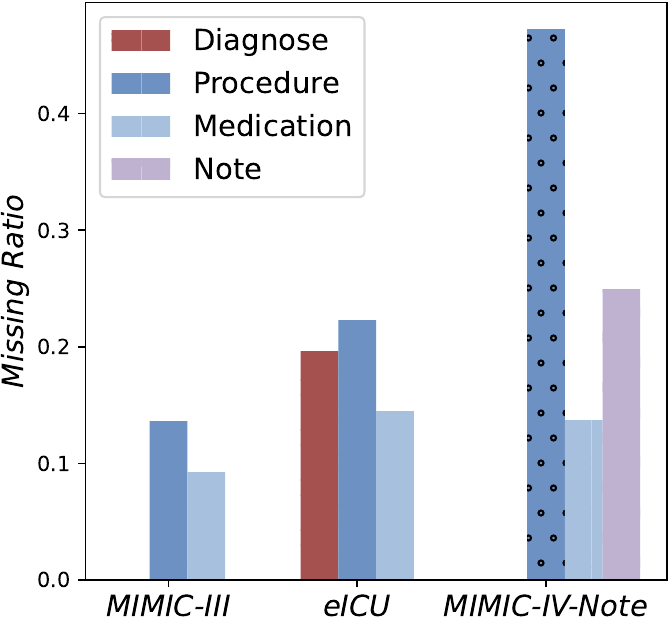}
\label{fig:motiv:miss}
\end{minipage}%
}%
\subfigure[PHE Pred (MIMIC-III)]{ 
\begin{minipage}[t]{0.32\linewidth}
\centering
\includegraphics[width=\linewidth,height=0.62\linewidth]{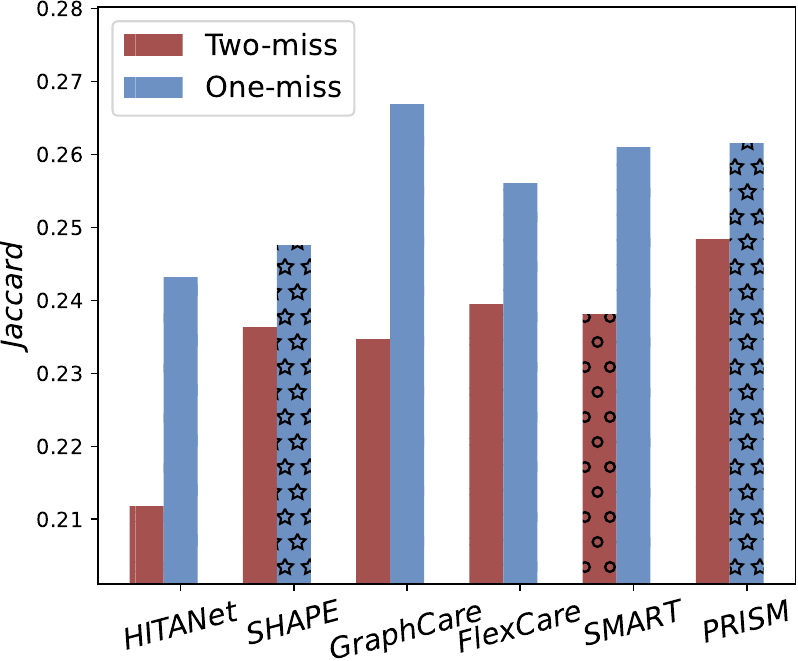}
\label{fig:motiv:per}
\end{minipage}
}%
   \caption{(a)-(b) refer to random missing view and missing status across three datasets. (c) Phenotype (PHE) prediction for patients with at most one view missing (One-miss) and two views missing (Two-miss).} 
   \label{fig:motiv}
   \vspace{-1em}
\end{figure}

Despite their effectiveness, these methods typically assume that views of EHRs for training are complete and that models can evenly utilize various views~\cite{yang2023molerec,yang2023pyhealth,yao2024drfuse}—two assumptions that are often unrealistic in real-world scenarios. For instance, regarding view missingness, existing popular EHR datasets~\cite{johnson2016mimic,johnson2023mimic,pollard2018eicu} frequently fail to provide complete diagnosis, procedure, and medication data for every patient visit due to issues like technical failures, incomplete data entry or variations in healthcare provider practices, as evidenced in Figure~\ref{fig:motiv:missing}-\ref{fig:motiv:miss}~\footnote{See Appendix~\ref{app:diff} for differences in missing definitions.}.  Current approaches~\cite{yang2023molerec,yang2023pyhealth,xu2023seqcare} often filter out these incomplete visits, leading to significant information loss and potentially skewed predictions. As shown in Figure~\ref{fig:motiv:per}, advanced methods demonstrate superior performance for patients with fewer missing EHR data, while performance declines significantly for those with more missing views. A prevalent approach is to identify transformation among various views of EHR data and map the observable views to impute the missing ones~\cite{zhang2022m3care,wu2024comprehensive}. However, pairwise measures typically necessitate training multiple mapping functions to accommodate the diverse contexts of random missingness, leading to computational inefficiencies and potentially compromising overall performance~\cite{li2023generating,zhong2024synthesizing}. For instance, with four views of EHR data, there are 15 potential missing patterns ($2^{4}-1$, excluding the complete view), requiring the training of 15 distinct mappings. Furthermore, these methods neglect the necessity of maintaining intra- and inter-patient consistency in reconstructed representations. Intra-patient consistency ensures coherence in an individual's historical data, while inter-patient consistency reflects trends among similar patients—both crucial for effective diagnostics~\cite{jiang2023graphcare,zhao2024enhancing}. For instance, a favorable response to treatment in one patient with diabetes may follow their history of insulin sensitivity or suggest similar outcomes for others with the same complications.

In addition to random missing views, view laziness presents another significant challenge. View laziness refers to a phenomenon where a model inadequately utilizes or ignores certain views or modalities, relying predominantly on a subset of available information, which can lead to suboptimal performance and diminished generalization~\cite{han2024erl}. Evidence in Figure~\ref{fig:case:inverse} indicates that, without constraints, existing approaches often encounter a scenario where a dominant data type—such as diagnosis—disproportionately influences model predictions, overshadowing insights from less dominant views, like medication history~\cite{xu2024flexcare}. This tendency to favor a certain view arises from a tendency to take shortcuts during model training, as stochastic gradient optimization prioritizes data views that have larger gradients in the short term~\cite{zhu2024prism,han2024erl}. Such an approach can lead to an imbalanced utilization of different data views, resulting in models that fail to leverage the full spectrum of available information~\cite{lee2023learning,yao2024drfuse}. For instance, neglecting historical medication data may overlook critical patterns in patient responses to treatments, which could inform more effective predictions. We also confirm the negative impact of view laziness on predictive performance from an empirical perspective in Table~\ref{tab:view:la}.

To address these challenges, we propose \textbf{Diffmv}, an innovative \textbf{diff}usion-based approach aimed at achieving better exploitation of \textbf{m}ulti-\textbf{v}iew EHR data. Specifically, first, to address the diversity of random missingness, we consolidate the generation processes of various EHR views within a unified diffusion-denoising pipeline, supplemented by a binary mask matrix to guide the model in generating the missing views. This progressive approach enables the model to learn heterogeneous relationships between views during the noise-adding and denoising stages. The incorporation of a binary mask matrix is crucial, as it facilitates the model’s understanding of any transformation between observable and missing views, thereby mitigating the computational burden associated with training multiple pairwise mapping functions. This not only enhances efficiency but also ensures that the model can better capture the complexities of inter-view relationships. Further, to ensure consistency between the generated views and the real-world treatment, we incorporate two contextual conditions—historical representations and prototype representations—during the denoising process. This approach mimics the real diagnostic process of healthcare professionals, who consider both the longitudinal health trajectories of individual patients and the collaborative treatment context involving similar patients. By integrating these conditions into the generation process, we improve generation fidelity, producing plausibly realistic EHR data for missing views, as evidenced in Figure~\ref{fig:case}. Ultimately, we leverage the imputed complete EHR data to foster more reliable healthcare outcomes. Second, to mitigate view laziness, we draw upon reinforcement learning~\cite{arulkumaran2017deep} by introducing the concept of relative advantage. This approach assesses the contributions of different views relative to the fused representation during optimization, allowing us to apply inverse advantage reweighting. For instance, if the view of clinical notes tends to dominate the predictions, this method will reduce their weight while increasing the weight of less dominant views, such as procedures or medication histories. This strategy promotes balanced utilization of all data views, preventing the model from relying solely on partial information for inference.

To summarize, our contributions are as follows:
\vspace{-0.1em}
\begin{itemize}[leftmargin=12pt]
    \item To the best of our knowledge, we are the first to simultaneously tackle two critical challenges in utilizing multi-view EHRs: random missing views and view laziness. Our proposed Diffmv represents an innovative generative solution.

    \item We uniquely consolidate multi-view generation within a unified diffusion-denoise framework, using multiple conditions to enhance imputation quality. Our novel reweighting loss optimizes view utilization, ensuring cohesive integration of all views.

    \item Extensive comparative experiments and robustness tests across multiple healthcare prediction tasks on three datasets in both multi-view and multi-modality scenarios demonstrate our framework's superiority. Our code is publicly available on GitHub~\footnote{https://github.com/xmed-lab/MMHealth}.
\end{itemize}

\section{Related Work}\label{sec:rel}
In this section, we review the closely related work, highlighting both connections and distinctions.

\subsection{Healthcare Prediction}\label{sec:2.1}
Healthcare prediction refers to the use of data-driven methodologies to forecast patient outcomes and optimize resource allocation~\cite{gao2024compositional}. This encompasses various applications, such as phenotype prediction~\cite{yu2024smart} and length-of-stay prediction~\cite{yang2023pyhealth}.

Existing literature on healthcare prediction can be broadly categorized into three genres~\cite{yang2023pyhealth,wang2024recent}: rule-based, instance-based, and longitudinal-based approaches. Rule-based methods leverage a set of predefined rules established by experts to derive predictions, e.g. MetaMap system~\cite{berge2023combining}. While this approach can yield interpretable results, it often struggles with scalability and adaptability to complex clinical scenarios. In contrast, instance-based paradigms focus on the current representations of patients, utilizing mechanisms such as attention or graph neural networks to extract relationships between EHR entities. For example, the LEAP~\cite{zhang2017leap} employs content-based attention to capture mappings between disease and medication labels. This data-driven approach, however, tends to overlook the rich information embedded in historical disease progression due to its static perspective. To address these limitations, longitudinal-based frameworks have emerged~\cite{zhao2024enhancing,jiang2023graphcare}, framing healthcare prediction as a sequential modeling challenge. Notably, models like DEPOT~\cite{zhao2024enhancing} and SHAPE~\cite{liu2023shape} utilize LSTM and Transformer architectures, respectively, to integrate prior knowledge within healthcare contexts for enhanced predictions. These longitudinal methods demonstrate significant advantages over their rule-based and instance-based counterparts. Despite their effectiveness, most studies assume complete EHR views and that models can uniformly leverage various data views. A few studies, such as M3Care~\cite{zhang2022m3care} and SMART~\cite{yu2024smart}, are designed to address specific view deficiencies from mapping and pre-training perspectives, thus enhancing downstream predictions.

\textit{In contrast to prior literature, our work addresses two critical issues—random missing views and view laziness—to enhance healthcare prediction performance. While M3Care and SMART explore specific view deficiencies, they inadequately tackle the complexities of random missing views and fail to investigate intra- and inter-patient consistency thoroughly. Additionally, neither approach sufficiently addresses the challenges of view laziness.}


\subsection{Incomplete Multi-view Learning}\label{sec:2.2}
Multi-view datasets often suffer from incompleteness due to equipment failures or human oversight. This phenomenon leads to the development of incomplete multi-view learning (IML)~\cite{zhou2024survey}.

Current strategies targeting IML can be technically categorized into discriminative approaches and generative approaches~\cite{wu2024comprehensive}. Discriminative methods optimize the target function based on assumptions of consistency and complementarity, seeking to identify similarities and shared representations among multi-view data. Common paradigms include mapping function~\cite{zhang2022m3care,zhu2024prism}, matrix factorization~\cite{lv2023joint} and graph completion~\cite{liu2023self}. While these methods can effectively leverage existing data to infer missing information, they may struggle with scalability and may not fully capture the underlying data distribution. In contrast, generative approaches focus on developing models that learn the inherent distribution of multi-view data. This category can be further divided into AutoEncoder-based~\cite{liu2023m3ae}, GAN-based~\cite{bernardini2023novel}, and contrastive learning-based methods~\cite{qiao2024multi}. Recently, the rise of diffusion-based models prompts researchers to explore their application in the progressive generation of missing views~\cite{zhou2024survey,meng2024multi,ruan2023mm}. cCGAN~\cite{bernardini2023novel} and MedDiffusion~\cite{zhong2024meddiffusion} serve as generative representatives in the field of healthcare.

\textit{Our model aligns with the generative approach, specifically leveraging diffusion techniques to generate missing views in healthcare scenario. Unlike discriminative IML, our method employs a progressive completion strategy that sharpens the learning of heterogeneous relationships between views. In contrast to existing diffusion-based IMLs that utilize multiple diffusion-denoising processes for different views, our approach consolidates multiple views into a unified diffusion-denoising framework, augmented by various conditions to facilitate classifier-free generation. This results in a more efficient and effective strategy for addressing the challenges of incomplete multi-view data.}

\section{Proposed Method}\label{sec:method}

\noindent\textbf{Preliminary.}
Each patient's medical history is represented as a sequence of visits, $\mathcal{U}^{(i)} = (\mathbf{u}^{(i)}_{1}, \mathbf{u}^{(i)}_{2}, \dots, \mathbf{u}^{(i)}_{\mathcal{T}_{i}})$, where $i$ identifies the $i$-th patient in the set $\mathcal{U}$ and $\mathcal{T}_{i}$ indicates the total number of visits. Each visit yields multiple views of data, represented as $\mathbf{u}_{j}^{(i)} = \{\mathbf{v}^{d}_{j}, \emptyset, \ldots, \mathbf{v}^{m}_{j}\}$, where some views may be missing, denoted by \( \emptyset \).
These views typically include diagnosis (\({d} \in \mathcal{D}\)), procedure (\({p} \in \mathcal{P}\)), and medication (\({m} \in \mathcal{M}\)), each represented as  multi-hot vectors $\mathrm{d}, \mathrm{p}, \text{and} \ \mathrm{m}$ repectively~\footnote{MIMIC-III and eICU contain three EHR views: $d$, $p$, and $m$. MIMIC-IV-Note includes $d$, $p$, $m$, and clinical notes ($cn$), providing an additional view for analysis.}. For instance, in the \(j\)-th visit \(\{\mathbf{v}^{{d}}_{j}, \emptyset, \emptyset\}\), the patient has a diagnosis represented by the vector \({\mathrm{d}}_{j} = [1, 1, 0, 0]\), indicating two diseases (where \(|\mathcal{D}| = 4\)), while the procedure and medication views are missing. Our goal is to impute the missing views $\emptyset$ and balance the utilization of all views to improve prediction accuracy.
For clarity, the patient index $i$ is omitted in the following content.

\noindent\textbf{Task formulation.} Following~\cite{jiang2023graphcare,yang2023pyhealth,xu2024flexcare}, we outline the definitions of the two common healthcare prediction tasks. 

\begin{itemize}[leftmargin=12pt]
\vspace{-0.3em}
 \item \textbf{Phenotype Prediction (PHE Pred)} involves a multi-label classification task aimed at identifying potential phenotypes based on patient data. This task focuses on analyzing $[\mathbf{u}_{1}, ..., \mathbf{u}_{j}]$ to predict the phenotype set $\mathrm{g}_{j+1}$ at time $j+1$, where the target $\mathbf{y}[\mathbf{u}_{j+1}] \in \mathbb{R}^{1\times |\mathcal{G}|}$, with $\mathrm{g}_{j+1} \in \mathcal{G}$. $\mathcal{G}$ is phenotype set~\cite{moseley2020phenotype,yu2024smart}.

\item \textbf{Length of Stay Prediction (LOS Pred)} involves  multi-classif-\\ication of forecasting the patient's hospitalization duration, informed by \(\{\mathbf{u}_{1}, ..., \mathbf{u}_{j+1}\}\). The duration, \(\mathcal{C}\), are divided into 10 splits like~\cite{jiang2023graphcare,yang2023pyhealth}, with \(\mathbf{y}[\mathbf{u}_{j+1}]\) denoting the category in \(\mathbb{R}^{1\times |\mathcal{C}|}\).
\end{itemize}

\begin{figure*}[!h]
\centering
\includegraphics[width=0.98\linewidth,height=0.32\linewidth]{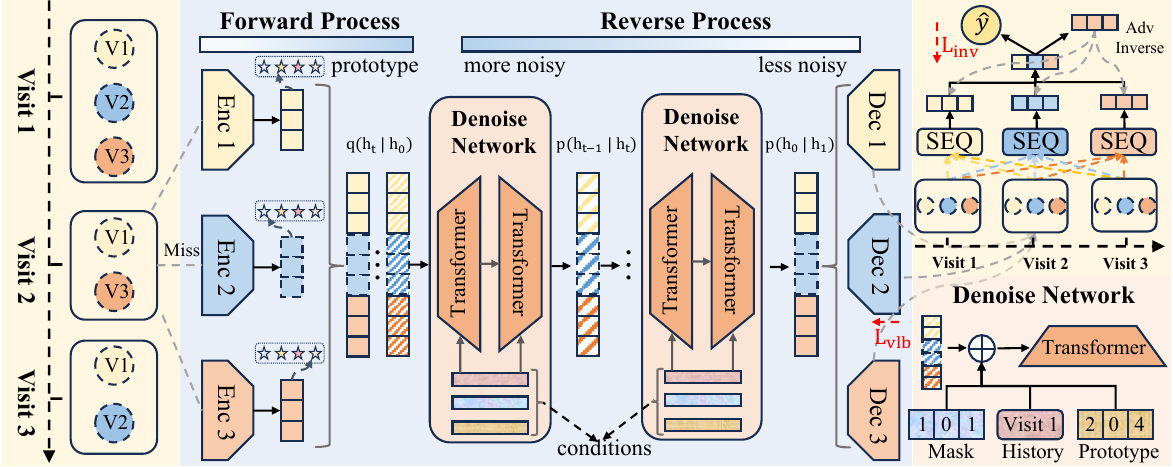}
\centering
\setlength{\abovecaptionskip}{-0.01cm}   
\caption{Overview of \textit{Diffmv}. We first vectorize the EHR data for each view (V) and apply forward diffusion to generate noisy vectors. These vectors are then concatenated and fed into a unified denoising network for reverse denoising, using a binary mask, historical data, and prototypes as conditions. Next, we utilize the trained diffusion framework to impute missing views and pass the EHRs to a sequential model (SEQ), followed by inverse advantage (Adv) reweighting to refine predictions. "Enc" refers to visit encoding in Eq.\ref{eq:1}-\ref{eq:2}, while "Dec" denotes the rounding process in Eq.~\ref{eq:9}. Both $q$ and $p$ signify conditional probability distributions, distinguishing between the forward and reverse processes. $\mathcal{L}_{\text{vlb}}$ and $\mathcal{L}_{\text{inv}}$ refer to Eq.~\ref{eq:14} and Eq.~\ref{eq:18}.}
\label{fig:frame}
\vspace{-1em}
\end{figure*}

\noindent\textbf{Solution Overview.}
Our solution targets the challenges of random missing views and view laziness in multi-view healthcare prediction, structured around two main processes: generative imputation and reweighting. First, we propose a unified diffusion-denoising framework that incorporates a binary mask matrix to adaptively model heterogeneous transformation patterns between different views. During the denoising process, we introduce multiple contextual conditions to ensure intra- and inter-patient consistency in the reconstructions. This approach mimics the real diagnostic process, enhancing the reliability of generated data for subsequent analyses.
Upon completing the training, we utilize the learned framework to generate missing views across the dataset, subsequently feeding the imputed data into any longitudinal-based model for patient representation extraction. Finally, to mitigate the model's potential over-reliance on the dominant view, we implement an inverse advantage reweighting scheme that rebalances the contributions of different view representations. This ensures a more equitable utilization of diverse sources, ultimately improving the robustness of the predictive outcomes. Our framework is illustrated in Figure~\ref{fig:frame}.

\subsection{Unified Diffusion Process}\label{sec:3.1}
Diffusions are likelihood-based generative models that learn data distributions through forward and reverse Markov processes~\cite{lu2022dpm}. Our goal is to develop a unified diffusion model that progressively learns the heterogeneous relationships among multiple views, thereby effectively imputing missing information. Following~\cite{yu2024smart, zhou2024survey, liu2023m3ae}, we employ complete EHRs from the dataset with random masks for training and perform imputation on the full dataset.

\noindent\textbf{Visit Encoder.}
The diffusion model was originally designed for continuous spaces, such as images, and cannot be directly applied to the generation of discrete codes in healthcare scenarios~\cite{li2022diffusion,bao2023all}. To adapt a continuous diffusion model for use in discrete spaces, we first employ embedding like~\cite{li2022diffusion} to project the discrete codes into the latent continuous space. Formally,
\begin{equation}\label{eq:1}
\setlength\abovedisplayskip{2pt}
\mathbf{e}^{d} = \mathbf{E}_{\mathcal{D}}({d}), \quad
\mathbf{e}^{p} = \mathbf{E}_{\mathcal{P}}({p}), \quad
\mathbf{e}^{m} = \mathbf{E}_{\mathcal{M}}({m}), 
\end{equation}
where $\mathbf{E}$ denotes the embedding function.
Next, we apply widely-used sum pooling~\cite{choi2016retain,zhao2024enhancing} to obtain visit-level representations. 
Formally, for $j$-th visit, 
\begin{equation}\label{eq:2}
\mathbf{v}^{{d}}_{j} = \sum_{d \in \mathrm{d}_{j}} \mathbf{e}^{d},\quad \mathbf{v}^{{p}}_{j} = \sum_{p \in \mathrm{p}_{j}} \mathbf{e}^{p}
,\quad \mathbf{v}^{{m}}_{j} = \sum_{m \in \mathrm{m}_{j}} \mathbf{e}^{m},
\end{equation}
where $\mathbf{v}^{*}$ denotes the visit-level embedding for view $*$. Note that for any missing view, $\mathbf{v}^{*}$ is zero vector.

\noindent\textbf{Forward Diffusion Process.}
To better capture the heterogeneous relationships among the views, we concatenate all views into a unified diffusion input. This approach enables the model to address any missing views within a single network and enhances generation by establishing an effective common latent space. Formally,
\begin{equation}\label{eq:3}
    \mathbf{v}_{j} = \mathbf{v}^{d}_{j} \oplus \mathbf{v}^{p}_{j} \oplus \mathbf{v}^{m}_{j},
\end{equation}
where $\oplus$ is a concatenation operation.
To address the challenges posed by arbitrary missing views, we innovatively incorporate a binary mask matrix $\mathbf{b}_{j} \in \mathbb{R}^{3*\tilde{d}}$ to identify any absent view, where $\tilde{d}$ refers to the embedding size. Formally, for each bit of $\mathbf{b}_{j}$,
\begin{equation}\label{eq:4}
b^{*}_{j}= \begin{cases}\mathbf{E}_{\mathcal{B}}(1) & \text { if } \mathbf{v}^{*}_{j} \text { is observed } \\ \mathbf{E}_{\mathcal{B}}(0) & \text { if } \mathbf{v}^{*}_{j} \text { is missing }\end{cases},
\end{equation}
where $\mathbf{E}_{\mathcal{B}}$ is the mask embedding function. We treat this mask as a condition and concatenate it with the visit representation as the initial state of diffusion, that is, \(\mathbf{h}_{0} = \mathbf{b}_{j} \oplus \mathbf{v}_{j}\).
Next, we add noise to the missing views using a general Gaussian distribution, akin to the noise addition process described in~\cite{li2022diffusion,rombach2022high}. Specifically, for each forward step \( q(\mathbf{h}_t \ | \  \mathbf{h}_{t-1}) \), we gradually inject noise into the hidden state from the previous step, \( \mathbf{h}_{t-1} \), to obtain \( \mathbf{h}_t \). Formally, 
\begin{equation}\label{eq:5}
\mathbf{h}_t^{\prime} \sim q(\mathbf{h}_t \mid \mathbf{h}_{t-1})=\mathcal{N}(\mathbf{h}_t ; \sqrt{1-\beta_t} \mathbf{h}_{t-1}, \beta_t \mathbf{I}),
\end{equation}
where $\mathcal{N}$ refers to the Gaussian distribution and $\beta_{t}$ refers to the amount of noise added in the t-th step.
It is important to note that introducing noise to the missing views emphasizes the reconstruction relationship between the observable views and the missing views during the training phase.

\noindent\textbf{Reverse Denoising Process.}
In the reverse process, our objective is to recover the original representation from the contaminated noisy version. A common strategy involves sampling $t$ from a Gaussian distribution and then inputting it into the denoising network to predict the added noise~\cite{li2022diffusion,rombach2022high}. Formally,
\begin{equation}\label{eq:6}
\mathbf{h}_{t-1}^{\prime} \sim p(\mathbf{h}_{t-1} \mid \mathbf{h}_{t})=\mathcal{N}(\mathbf{h}_{t-1} ; \mu_\theta(\mathbf{h}_{t}, t), \sigma_\theta(\mathbf{h}_t, t)),
\end{equation}
where \(\theta\) refers to the denoising network. Leveraging the strong performance of DiT~\cite{peebles2023scalable} and the nature of sequential scenes, we select the Transformer as the denoising network. \(\mu\) and \(\sigma\) represent the estimated mean and variance using $\theta$, respectively.
Once the corresponding noise addition and denoising measures are established, we can approximate the variational lower bound for optimization like~\cite{peebles2023scalable}. Formally,
\begin{equation}\label{eq:7}\small
\begin{aligned}
\underset{\theta}{\mathcal{L}_{\text{vlb}}(\mathbf{h}_0,t)}&=\underset{q(\mathbf{h}_{1: T} \mid \mathbf{h}_0)}{\mathbb{E}}[\log \frac{q(\mathbf{h}_T \mid \mathbf{h}_0)}{p_\theta(\mathbf{h}_T)}+\sum_{t=2}^T \log \frac{q(\mathbf{h}_{t-1} \mid \mathbf{h}_0, \mathbf{h}_t)}{p_\theta(\mathbf{h}_{t-1} \mid \mathbf{h}_t)}\\
& -\log p_\theta(\mathbf{h}_0 \mid \mathbf{h}_1)],
\end{aligned}
\end{equation}
where $\mathbb{E}$ denotes mathematic expectation and $T$ refers to the total diffusion steps.
However, this approach is only applicable in continuous space, making it challenging to generate discrete targets, such as specific medical codes in our scenario. To address this, we mimic the autoregressive process and introduce a rounding $p_{\theta}(\mathbf{w}\ |\ \mathbf{h})$ that transforms continuous space into discrete space. Formally,
\begin{equation}\label{eq:8}\small
\begin{aligned}
\underset{\theta}{\mathcal{L}_{\text{vlb}}(\mathbf{h}_{0},t)} & =\underset{q(\mathbf{h}_{1: T} \mid \mathbf{h}_0)}{\mathbb{E}}[\log \frac{q(\mathbf{h}_T \mid \mathbf{h}_0)}{p_\theta(\mathbf{h}_T)}+\sum_{t=2}^T \log \frac{q(\mathbf{h}_{t-1} \mid \mathbf{h}_0, \mathbf{h}_t)}{p_\theta(\mathbf{h}_{t-1} \mid \mathbf{h}_t)} \\
& + \underbrace{\log \frac{q_\phi(\mathbf{h}_0 \mid \mathbf{w}^{d \oplus p \oplus m})}{p_\theta(\mathbf{h}_0 \mid \mathbf{h}_1)}}_{\mathcal{L}_{\text{emb}}}-\underbrace{\log p_\theta(\mathbf{w}^{d \oplus p \oplus m} \mid \mathbf{h}_0)]}_{\mathcal{L}_{\text{round}}},
\end{aligned}
\end{equation}
where $p_{\theta}(\mathbf{w} |\ \mathbf{h})$ denotes the probability of the  medical entities given $\mathbf{h}$, akin to a classification task. $q_{\phi}(\mathbf{h}\ |\ \mathbf{w})$ refers to the embedding process in Eq.~\ref{eq:1}-\ref{eq:3}, where $\phi$ can be regarded as the embedding layer of $\theta$.
We further simplify the training objective according to the parameterization trick used in~\cite{sohl2015deep}. Formally,
\begin{equation}\label{eq:9}\small
\begin{aligned}
\underset{\theta}{\mathcal{L}_{\text{vlb}}(\mathbf{h}_{0},t)}  & = [\sum_{t=2}^T||\mathbf{h}_0-g_\theta(\mathbf{h}_t, t)||^2+||{\text{EMB}}(\mathbf{w}^{d \oplus p \oplus m})-g_\theta(\mathbf{h}_1, 1)||^2\\
&-\log p_\theta(\mathbf{w}^{d \oplus p \oplus m} \mid \mathbf{h}_0)], \\
\end{aligned}
\end{equation}
where $\text{EMB}(\cdot)$ denotes the embedding function in Eq.~\ref{eq:1}-~\ref{eq:3}. $g_\theta(\cdot)$ refers to the denoising network Transformer. In this way, we innovatively integrate continuous and discrete spaces into the denoising process, compelling the model to focus on the semantics of discrete entities while effectively reconstructing the target vector. More detailed theoretical derivation for the forward and reverse processes can be found in Appendix~\ref{app:theory}.

\subsection{Consistency-aware Encoder}\label{sec:3.2}
In the previous section, we aim to exploit the heterogeneous relationships among medical entities within the same visit to enhance the reconstruction of missing views. Considering the "temporal" and "relational" characteristics that are essential in healthcare diagnostic scenarios—specifically, the importance of a patient's historical conditions over time and their relationships with similar patients—we refine our approach~\cite{jiang2023graphcare,zhao2024enhancing}. In the reverse phase, we integrate intra- and inter-patient consistency as conditions $\mathbf{C}$ to provide rich context, facilitating more nuanced estimates of patients' missing information, i.e., $p(\mathbf{h}_{t-1} \ | \ \mathbf{h}_{t}) \rightarrow p(\mathbf{h}_{t-1} \ | \ \mathbf{h}_{t} \oplus \mathbf{C}_{t})$.

\noindent\textbf{Intra-patient Consistency Encoder.}
In the typical diagnosis and treatment process, doctors frequently make decisions based on a patient's historical records. For example, a physician may review a patient's past medical history of hypertension and diabetes when assessing their risk for cardiovascular disease. Furthermore, research in ~\cite{xu2024flexcare,jiang2023graphcare} has demonstrated that incorporating longitudinal historical information can significantly enhance the model's predictive performance. This potential for intra-patient consistency inspires us to explore further improvements in generation quality. During the denoising process, we encode the patient's historical records and integrate them as conditional inputs. Formally,
\begin{equation}\label{eq:10}
\mathbf{z}_{j}=\operatorname{Avg}(\operatorname{softmax}(\frac{\mathbf{Q} \mathbf{K}^T}{\sqrt{\tilde{d}}}) \cdot \mathbf{V}_{1: j-1}),
\end{equation}
where
\(\mathbf{Q} = \mathbf{v}_{1:j-1} \mathbf{W}^Q\) represents the queries, 
\(\mathbf{K} = \mathbf{v}_{1:j-1} \mathbf{W}^K\) represents the keys, 
\(\mathbf{V}_{1:j-1} = \mathbf{v}_{1:j-1} \mathbf{W}^V\) represents the values for the visit embeddings from $[1,j-1]$. $\mathbf{W}^{Q,K,V} \in \mathbb{R}^{|\tilde{d} \times \tilde{d}|}$ refers to the shared transformation matrix.

\noindent\textbf{Inter-patient Consistency Encoder.}
In addition to reviewing a patient's historical health status, doctors also consider the diagnostic processes of similar patients when making a diagnosis. For example, a doctor might reference the treatment outcomes of patients with similar medical history, and symptoms to inform their decision~\cite{zhang2021grasp}. This approach is equally important in model generation, particularly for users with too few visits, who may lack rich historical records and thus cannot generate diverse views. To address this, we employ a prototype-based recall of similar patients' visits to enhance our generation capabilities. Formally, we first apply K-means~\cite{xu2024protomix} to obtain prototypes of visits,
\begin{equation}\label{eq:11}
\min _{\left\{\mathbf{c}^{*}_k\right\}} \sum_{k=1}^K \sum_{j=1}^N \mathbb{I}(c_j^{*}=k)||\mathbf{v}^{*}_j-\mathbf{c}^{*}_k||^2,
\end{equation}
where $\mathbb{I}$ is the indicator function. $K$ and $N$ refer to the total number of prototypes and visits, respectively. $c$ and $\mathbf{c}$ denote the prototype index and its  representation. Once the cluster prototypes are obtained, we can determine the group to which each sample belongs and use the prototype of that group as a coarse-grained representation of similar patients. Formally,
\begin{equation}\label{eq:12}
c_j^{*}=\arg \min _k||\mathbf{v}^{*}_j-\mathbf{c}_k^{*}||.
\vspace{-1em}
\end{equation}
This allows us to obtain a visit-level central representation, i.e., $\mathbf{c}_j=\mathbf{c}_j^{d} \oplus \mathbf{c}_j^{p} \oplus \mathbf{c}_j^{m}$. Note that when a view is missing, its corresponding prototype is padded with a vector of zeros.

\noindent\textbf{Classifier-free Approach.}
We then feed these two conditions into the denoising process in a concatenated manner as the auxiliary condition. Formally, we reformulate $\mathbf{h}_{0}$ in Eq.~\ref{eq:5} and Eq.~\ref{eq:9} as, 
\begin{equation}\label{eq:13}
\setlength\abovedisplayskip{2pt}
    {\mathbf{h}}_{0} = \underbrace{\mathbf{z}_{j} \ \oplus \ \mathbf{c}_{j} \ \oplus \ \mathbf{b}_{j}}_{\text{condition} \ \mathbf{C}_{0}} \oplus \ \mathbf{v}_{j},
\end{equation}
where the first two items can be considered as contextual conditions that support the final prediction.
We adopt a classifier-free guidance~\cite{ho2021classifier} to optimize the model, ensuring that it can effectively incorporate conditional information. Formally,
\begin{equation}\label{eq:14}\small
\underset{\theta}{\tilde{\mathcal{L}}_{\text{vlb}}(\mathbf{h}_t, \mathbf{C}_{t}, t)}=\underset{\theta}{\mathcal{L}_{\text{vlb}}(\mathbf{h}_t, \emptyset, t)}+s \cdot[\underset{\theta}{\mathcal{L}_{\text{vlb}}(\mathbf{h}_t, \mathbf{C}_{t}, t)}-\underset{\theta}{\mathcal{L}_{\text{vlb}}(\mathbf{h}_t, \emptyset, t)]},
\end{equation}
where $\emptyset$ entails replacing the condition with a zero vector of the same shape. $s$ signifies the guidance scale. This paradigm implicitly incorporates the gradients of conditions into the diffusion-denoise process, eliminating the need for additional explicit classifiers~\cite{ho2021classifier}.

\subsection{Longitudinal Patient Representation}\label{sec:3.3}
Once the diffusion model training is complete, we can input all incomplete visit-level representations from the training set to fill in the missing view. Specifically, we begin by randomly sampling \( t \sim \mathcal{N}(0, \mathbf{I}) \) and using Eq.~\ref{eq:5} to obtain \(\mathbf{h}^{\prime}_{t}\). We then repeat the denoising process outlined in Eq.~\ref{eq:6} until we reach \(\mathbf{h}_{0}\), and subsequently use \( p_{\theta}(\mathbf{w}\ |\ \mathbf{h}_{0}) \) to derive the discrete medical code. Formally, for each visit $j$ for any given patient, assuming that $p$ view is missing, 
\begin{equation}\label{eq:15}
   \mathbf{d}_{j}, \mathbf{p}_{j}, \mathbf{m}_{j} = \text{UDP}[{(\mathbf{d}_{j}, \emptyset, \mathbf{m}_{j}); \mathbf{b}_{j}, \mathbf{z}_{j},\mathbf{c}_{j}}],
\end{equation}
where UDP denotes the denoising process.
This enables us to construct a complete EHR for every visit of each patient. We can then apply any sequential model  $\mathcal{E}(\cdot)$ to extract temporal patterns. Formally, for each view,
\begin{equation}\label{eq:16}
\mathbf{u}^{d}_{j} = \mathcal{E}(\mathbf{d}_{1:j}), \mathbf{u}^{p}_{j} = \mathcal{E}(\mathbf{p}_{1:j}), \mathbf{u}^{m}_{j} = \mathcal{E}(\mathbf{m}_{1:j}). \\
\end{equation}
We use a Transformer for $\mathcal{E}(\cdot)$ as outlined in ~\cite{jiang2023graphcare,xu2024flexcare}, and conduct robust analysis in Appendix~\ref{sec:4.3.4}. Once the representations of each view are obtained, the traditional approach is to concatenate them for final healthcare prediction, i.e., $\hat{\mathbf{y}}_{j+1}=\mathbf{W}[\mathbf{u}^{d}_{j}\oplus\mathbf{u}^{p}_{j}\oplus\mathbf{u}^{m}_{j}]$, where $\mathbf{W} \in \mathbb{R}^{|3\tilde{d}\times\mathcal{G}|}$ for PHE Pred and $\mathbf{W} \in \mathbb{R}^{|3\tilde{d}\times\mathcal{C}|}$ for LOS Pred.

\subsection{Re-weighting Strategy \&  Optimization}\label{sec:3.4}
In this subsection, we introduce the optimization strategy.

\noindent\textbf{Inverse Advantage Weight.}
As discussed in Section~\ref{sec:4.3.5}, concatenating multi-view information~\cite{jiang2023graphcare,gao2020stagenet} often leads to uneven utilization of each view, limiting the model's generalization capabilities. To address this, we draw on the advantage function from reinforcement learning~\cite{arulkumaran2017deep,baird1994reinforcement,sun2024redcore} to assess the relative advantage of each view against a unified view. This evaluation allows us to apply inverse advantage reweighting, ensuring a more balanced contribution from each view. Formally, for advantage $\mathbf{r}_{j} = [r^{d}_{j}, r^{p}_{j},{r}^{m}_{j}]$, 
\begin{equation}\label{eq:17}\small
    {r}^{*}_{j} = \frac{\mathcal{L}_{\text{ce}}(\hat{\mathbf{y}}_{j+1},\mathbf{y}_{j+1}) - \mathcal{L}_{\text{ce}}(\mathbf{W}^{*}\mathbf{u}^{*}_{j},\mathbf{y}_{j+1})}{\mathcal{L}_{\text{ce}}(\mathbf{W}\mathbf{u}_{j}, \mathbf{y}_{j+1})},
\end{equation}
where $\mathcal{L}_{\text{ce}}$ denotes the cross entropy loss for task prediction and $\mathbf{u}_{j}=\mathbf{u}^{d}_{j}\oplus\mathbf{u}^{p}_{j}\oplus\mathbf{u}^{m}_{j}$. $\mathbf{W}^{*}$ is the final projection layer for view $*$.
Once we have $\mathbf{r}$, we can adaptively perform inverse advantage reweighting. Let \( \mathbf{n} = [n^{d}, n^{p}, n^{m}] \) denote a learnable weight vector. To minimize the overall advantage—ensuring that all views demonstrate no relative advantage—the loss is formulated as follows,
\begin{equation}\label{eq:18}
\mathcal{L}_{\text{inv}}(\mathbf{r}, \mathbf{n}) = \mathbf{r}^{\mathbf{T}} \cdot \mathbf{n}, \quad \text{s.t.} \quad \sum_{*\in\{d,p,m\}} n^{*} = 1, \quad n^{*} > 0 \, \forall *,
\end{equation}
where we constrain the sum of 
$\mathbf{n}$ to be 1, with weights being positive numbers to ensure their semantic meaning.
This loss function directly constrains the relative advantages of each view, thereby explicitly promoting a balanced utilization.

\noindent\textbf{Total Loss.} 
Our total loss consists of two components: the weighted task loss and Eq.~\ref{eq:18}. Formally,
\begin{equation}\label{eq:19}
\mathcal{L}_{\text{total}} = \sum_{*\in\{d,p,m\}}n^{*}\mathcal{L}_{\text{ce}}(\mathbf{W}^{*}\mathbf{u}^{*}_{j}, \mathbf{y}_{j})+ \eta{\mathcal{L}_{\text{inv}}}(\mathbf{r}, \mathbf{n}),
\end{equation}
where $\eta$ is the balance weight.
Please note that the description of the method section is based on the three-view EHR datasets, MIMIC-III and eICU. In contrast, MIMIC-IV-Note comprises four views: d, p, m, and clinical notes (cn). The only difference is that for the note view, we utilize the widely used pre-trained language encoder, Sap-BERT~\cite{liu2021self}, to extract representations, represented as \( \mathbf{v}^{cn}_{j} = \mathbf{W}^{cn}(\text{Sap-BERT}(cn_{j})) \), with \( \mathbf{W}^{cn} \) ensuring dimensional consistency with other views. The remainder of the process is seamlessly integrated with this method. We also experimented with various pre-trained language models, as outlined in Appendix~\ref{sec:4.3.4}. Mathematical notations and a streamlined algorithmic flow are detailed in Appendix~\ref{app:math} and Algorithm~\ref{alg1}.
\vspace{-1em}
\begin{algorithm}\small 
\caption{The Algorithm of \textit{Diffmv}} 
\label{alg1} 
\begin{algorithmic}[1] 
\REQUIRE Multi-view EHRs $\mathcal{U}$, Guidance scale $s$;
\ENSURE Denoising network parameter $\theta$, longitudinal encoder parameter $\Theta$;
\STATE \textbf{Stage 1: Forward Diffusion} \Comment{No parameter training;}
\STATE Visit-level encoding and concatenation in Eq~\ref{eq:1}-\ref{eq:3} ; 
\STATE Forward diffusion adds noise to get $\mathbf{h}_t$ in Eq.~\ref{eq:5};
\STATE \textbf{Stage 2: Reverse Denoise}
\Comment{Tuning denoising network $\theta$}
\WHILE{not converged}
\STATE Sample t from Gaussian distribution;
\STATE Get intra-patient consistency condition in Eq.~\ref{eq:10};
\STATE Get inter-patient consistency condition in Eq.~\ref{eq:12};
\STATE Take classifier-free optimization in Eq.~\ref{eq:14};
\STATE Update the parameters;
\ENDWHILE
\STATE \textbf{Stage 3: Sampling and Training} \Comment{Tuning longitudinal encoder $\Theta$}
\STATE Sampling for all incomplete visits in $\mathcal{U}$ in Eq.~\ref{eq:15};
\WHILE{not converged}
\STATE View encoding in Eq.~\ref{eq:16};
\STATE Inverse advantage reweighting in Eq.~\ref{eq:17}-\ref{eq:18};
\STATE Optimization in Eq.~\ref{eq:19};
\STATE Update the parameters;
\ENDWHILE
\STATE \RETURN{} Parameters $\theta$ \& $\Theta$;
\end{algorithmic}
\end{algorithm}

\begin{table*}[!h]\small
\centering
\setlength{\abovecaptionskip}{-0.05cm}   
\setlength{\belowcaptionskip}{-0.1cm}   
\caption{Performance comparison: Phenotype Prediction. Bold indicates optimal performance.} %
\label{tab:phe}
\resizebox{0.92\textwidth}{!}{
\begin{tabular}{c|c|c|c|c||c|c|c|c||c|c|c|c} 
\toprule
Dataset       & \multicolumn{4}{c||}{MIMIC-III}    & \multicolumn{4}{c||}{eICU}     & \multicolumn{4}{c}{MIMIC-IV-Note}            \\ 
\hline
Method        & Jaccard & F1-score & AUPRC & AUROC & Jaccard & F1-score & AUPRC & AUROC & Jaccard & F1-score & AUPRC & AUROC  \\ 
\hline
GRASP~\cite{zhang2021grasp}         & 0.2088      & 0.4022       & 0.4297    & 0.8947    & 0.7966     & 0.9468       & 0.8978    & 0.9715         & 0.2767         & 0.4935      & 0.5159   & 0.9224         \\
RETAIN~\cite{choi2016retain}        & 0.2631      & 0.4346       & 0.4700    & 0.9025    & 0.7955     & 0.9412       & 0.8954    & 0.9722    & 0.2809         & 0.5035         & 0.5240      & 0.9207       \\
AdaCare~\cite{ma2020adacare}       & 0.2481     & 0.4330       & 0.4805    & 0.9070      & 0.8138     & 0.9592       & 0.8976    & 0.9719    & 0.2724         & 0.4806         & 0.4916      & 0.9129        \\
StageNet~\cite{gao2020stagenet}      & 0.2506      & 0.4309       & 0.4548    & 0.8974    & 0.8004     & 0.9412       & 0.8961    & 0.9700    & 0.2929         & 0.5125         & 0.5147      & 0.9198    \\
DEPOT~\cite{zhao2024enhancing}      & 0.2530      & 0.4361       & 0.4603    & 0.9004   & 0.8142      & 0.9528       & 0.8950    & 0.9718    & 0.2905         & 0.5073         & 0.5251      & 0.9245    \\
Transformer~\cite{vaswani2017attention} & 0.2372      & 0.4207       & 0.4596    & 0.9042    & 0.7819     & 0.9318       & 0.8841    & 0.9687   & 0.2838         & 0.5009         & 0.5208      & 0.9246   \\
HITANet~\cite{luo2020hitanet}  & 0.2446      & 0.4262       & 0.4672    & 0.9040    & 0.8064     & 0.9505       & 0.8911    & 0.9638      & 0.2876         & 0.5109         & 0.5078      & 0.9185   \\
SHAPE~\cite{liu2023shape}      & 0.2485      & 0.4235       & 0.4768    & 0.9067    & 0.8067     & 0.9315       & 0.8986    & 0.9706    & 0.2848         & 0.5004         & 0.5174      & 0.9228    \\
GraphCare~\cite{jiang2023graphcare}     & 0.2627      & 0.4383       & 0.4889    & 0.9099    & 0.8062     & 0.9399       & 0.8910    & 0.9705    & 0.2961         & 0.5126         & 0.5178      & 0.9209       \\
FlexCare~\cite{xu2024flexcare}      & 0.2629      & 0.4445       & 0.4864           & 0.9114    & 0.8161  & 0.9463         & 0.9032      & 0.9667     & 0.3031     & 0.5228    & 0.5307     & 0.9272        \\
SMART~\cite{yu2024smart}   & 0.2562         & 0.4306         & 0.4818      & 0.9094       & 0.8057     & 0.9429       & 0.8934    & 0.9702   & 0.2873      & 0.5035       & 0.5231    & 0.9232     \\
M3Care~\cite{zhang2022m3care}    & 0.2694         & 0.4488         & 0.4834      & 0.9108       & 0.8185   & 0.9575       & 0.8971    & 0.9703   & 0.2995      & 0.5210       & 0.5293    & 0.9261        \\
PRISM~\cite{zhu2024prism}      & 0.2616      & 0.4342       & 0.4794    & 0.9077     & 0.8199     & 0.9561       & 0.9046    & 0.9731    & 0.3003     & 0.5193       & 0.5299    & 0.9260      \\

cCGAN~\cite{bernardini2023novel}      & 0.2602      & 0.4385       & 0.4805    & 0.9063    & 0.7917     & 0.9318       & 0.8917    & 0.9727    & 0.2859         & 0.5063         & 0.5314      & 0.9205      \\
MedDiffusion~\cite{zhong2024meddiffusion}      & 0.2700      & 0.4446       & 0.4845    & 0.9135    & 0.8213     & 0.9640       & 0.8977    & 0.9689   & 0.2908         & 0.5154         & 0.5193      & 0.9246        \\
\hline
  \rowcolor[gray]{0.8}  \textit{Diffmv}  & \textbf{0.2831}      & \textbf{0.4534}       & \textbf{0.4902}    & \textbf{0.9139}    & \textbf{0.8348}     & \textbf{0.9662}       & \textbf{0.9108}    & \textbf{0.9761}   & \textbf{0.3144}         & \textbf{0.5248}         & \textbf{0.5321}      & \textbf{0.9283}\\
\bottomrule
\end{tabular}}
\vspace{-1em}
\end{table*}
\begin{table*}[!h]\small
\centering
\setlength{\abovecaptionskip}{-0.05cm}   
\setlength{\belowcaptionskip}{-0.1cm}   
\caption{Performance comparison: LOS Prediction.} %
\label{tab:los}
\resizebox{0.92\textwidth}{!}{
\begin{tabular}{c|c|c|c|c||c|c|c|c||c|c|c|c} 
\toprule
Dataset       & \multicolumn{4}{c||}{MIMIC-III}    & \multicolumn{4}{c||}{eICU}     & \multicolumn{4}{c}{MIMIC-IV-Note}            \\ 
\hline
Method        & Accuracy & F1-score & Kappa & AUROC & Accuracy & F1-score & Kappa & AUROC & Accuracy & F1-score & Kappa & AUROC  \\ 
\hline
GRASP~\cite{zhang2021grasp}         & 0.3467      & 0.3398       & 0.2421    & 0.7688    & 0.3693     & 0.3034       & 0.1250    & 0.6717   & 0.4148         & 0.3804         & 0.3039      & 0.8133       \\
RETAIN~\cite{choi2016retain}       & 0.3834      & 0.3485       & 0.2719    & 0.7903    & 0.3733     & 0.3009       & 0.1228    & 0.6743    & 0.4161         & 0.3854         & 0.3098      & 0.8147       \\
AdaCare~\cite{ma2020adacare}      & 0.3826      & 0.3413       & 0.2715    & 0.7961    & 0.3716     & 0.3090       & 0.1272    & 0.6688    & 0.4206         & 0.3888         & 0.3136      & 0.8180       \\
StageNet~\cite{gao2020stagenet}     & 0.3761      & 0.3202       & 0.2566    & 0.7899    & 0.3706     & 0.3155       & 0.1407    & 0.6754    & 0.4162         & 0.3868         & 0.3162      & 0.8160       \\
DEPOT~\cite{zhao2024enhancing}     & 0.3901      & 0.3537       & 0.2837    & 0.8015    & 0.3763     & 0.3049       & 0.1356    & 0.6767    & 0.4208         & 0.3901         & 0.3142      & 0.8198       \\
Transformer~\cite{vaswani2017attention}  & 0.4047      & 0.3709       & 0.3055    & 0.8186    & 0.3964     & 0.3419       & 0.1788    & 0.7232    & 0.3922         & 0.3517         & 0.2831      & 0.7976       \\
HITANet~\cite{luo2020hitanet} & 0.4091      & 0.3577       & 0.2993    & 0.8207    & 0.4026     & 0.3538       & 0.1904    & 0.7236     & 0.3923         & 0.3687         & 0.2839      & 0.7968       \\
SHAPE~\cite{liu2023shape}     & 0.4133      & 0.3576       & 0.3103    & 0.8261    & 0.4012     & 0.3511       & 0.1877    & 0.7238    & 0.4390         & 0.4291         & 0.3440      & 0.8389       \\
GraphCare~\cite{jiang2023graphcare}   & 0.4111      & 0.3862       & 0.3165    & 0.8275    & 0.4029     & 0.3459       & 0.1860    & 0.7200    & 0.4310         & 0.4023         & 0.3387      & 0.8276       \\
FlexCare~\cite{xu2024flexcare}      & 0.4160      & 0.3628       & 0.3051    & 0.8184  &0.4014 &0.3474 &0.1845 &0.7204    & 0.4396         & 0.4331         & 0.3466      & 0.8409       \\
SMART~\cite{yu2024smart}      & 0.4114       & 0.3658      & 0.3078       & 0.8209          & 0.3979     & 0.3345       & 0.1752    & 0.7126    & 0.4042         & 0.3934         & 0.3085      & 0.8036       \\
M3Care~\cite{zhang2022m3care}      & 0.4200      & 0.3764       & 0.3116    & 0.8267    & 0.4095     & 0.3693       & 0.1985    & 0.7293    & 0.4285         & 0.4045         & 0.3284      & 0.8249      \\
PRISM~\cite{zhu2024prism}     & 0.4252      & 0.3941       & 0.3275    & 0.8350      & 0.4070     & 0.3617       & 0.1991    & 0.7280    & 0.4406         & 0.4253         & 0.3453      & 0.8403      \\
cCGAN~\cite{bernardini2023novel}    & 0.4171      & 0.3729       & 0.3106    & 0.8182  & 0.4080     & 0.3584       & 0.1982    & 0.7296       & 0.3933         & 0.3594         & 0.2821      & 0.7905      \\
MedDiffusion~\cite{zhong2024meddiffusion}      & 0.4303      & 0.3741       & 0.3207    & 0.8368    & 0.4097     & 0.3536       & 0.1979    & 0.7319  & 0.4501         & 0.4251         & 0.3536      & 0.8450       \\
\hline
 \rowcolor[gray]{0.8} \textit{Diffmv}  & \textbf{0.4468}      & \textbf{0.4100}       & \textbf{0.3495}    & \textbf{0.8468}    & \textbf{0.4195}     & \textbf{0.3751}       & \textbf{0.2209}    & \textbf{0.7420}     & \textbf{0.4656}         & \textbf{0.4446}         & \textbf{0.3730}      & \textbf{0.8556}  \\
\bottomrule
\end{tabular}}
\vspace{-1em}
\end{table*}
\section{Experiments}\label{sec:exp}
We first outline the necessary setup and then present the analysis.
\subsection{Experimental Setup}\label{sec:4.1}

\noindent\textbf{Datasets \& Baselines.}
In this study, we conduct comprehensive experiments on three widely used datasets: MIMIC-III~\cite{johnson2016mimic}, eICU~\cite{pollard2018eicu}, and MIMIC-IV-Note~\cite{johnson2023mimic}. 
Their detailed statistics are provided in Appendix~\ref{app:data}. Please note that the corresponding statistics vary depending on the task definitions.

Our baselines encompass a range of well-established approaches, including 
GRASP~\cite{zhang2021grasp}, RETAIN~\cite{choi2016retain}, 
AdaCare~\cite{ma2020adacare}, StageNet~\cite{gao2020stagenet}, 
DEPOT~\cite{zhao2024enhancing}, 
Transformer~\cite{vaswani2017attention}, HITANet~\cite{luo2020hitanet}, SHAPE~\cite{liu2023shape}, GraphCare~\cite{jiang2023graphcare}, FlexCare~\cite{xu2024flexcare}, 
SMART~\cite{yu2024smart},
M3Care~\cite{zhang2022m3care},   PRISM~\cite{zhu2024prism},  cCGAN~\cite{bernardini2023novel} and MedDiffusion~\cite{zhong2024meddiffusion}.  Prior to GraphCare, algorithms primarily rely on concatenating various views of EHRs to identify high-order and sequential patterns, without specifically distinguishing between each view.
FlexCare and SMART are tailored for multi-view healthcare prediction. M3Care and PRISM address missing views by employing discriminative mapping for imputation, whereas cCGAN and MedDiffusion concentrate on synthesizing EHR data to enhance the original sequence.
For other algorithms without imputation capability, we utilize zero-vector to pad missing views.
This comprehensive selection of baselines enables a thorough evaluation of the superiority of our proposed method.

\noindent\textbf{Implementation Details \& Evaluations.}
We implement our algorithm using PyTorch, utilizing the Adam optimizer with a learning rate of 1e-3 and a weight decay of 5e-4. The batch size is set to 32, and we adopt an embedding size of 128 unless otherwise specified, consistent with~\cite{jiang2023graphcare,liu2023shape}. For both tasks, the training epochs are fixed at 50. In the diffusion, we configure the number of diffusions and the sampling steps to 1000 and 20, as established in~\cite{li2022diffusion}, and employ the DPM-solver~\cite{lu2022dpm} during the denoising phase to enhance computational efficiency. During inference, our computational cost is closely tied to the longitudinal encoder, incurring minimal additional burden.
We utilize the Transformer as our longitudinal feature encoder, with additional variations discussed in Appendix~\ref{sec:4.3.4}. The prototype number and the guidance scale in Eq.~\ref{eq:11}, in Eq.~\ref{eq:14} and balance weight in Eq.~\ref{eq:19} are set to 10, 0.5, and 1e-3, respectively. A comprehensive hyperparameter tuning for these critical parameters is detailed in Appendix~\ref{app:hyper}.

For data partitioning, we adhere to established practices~\cite{zhao2024enhancing} by dividing the datasets into training, validation, and test sets in a 6:2:2 ratio. For PHE Pred, we employ evaluation metrics including Jaccard, F1-score, AUPRC, and AUROC. In LOS Pred, we assess performance using Accuracy, F1-score, Kappa, and AUROC. These metrics are widely chosen for their clinical relevance and ability to provide a comprehensive assessment~\cite{jiang2023graphcare,yang2023pyhealth}. Their mathematical definitions are provided in Appendix~\ref{app:metric}. 
\vspace{-1em}
\subsection{Overall Performance}\label{sec:4.2}
In Table~\ref{tab:phe}-\ref{tab:los}, we present the performance of all algorithms, demonstrating that our proposed Diffmv achieves superior results. StageNet and RETAIN may not perform as well as Transformer-variants, potentially due to the cumulative noise introduced by missing views in the progressive addition of recurrent neural mechanism, while masked attention can partially address this issue. GraphCare also serves as a robust baseline, with a 3.5\% improvement compared to the DEPOT, benefiting from the incorporation of external knowledge.
FlexCare and SMART implement differentiated attention mechanisms tailored to various views of EHR data, resulting in more effective utilization. While both MedDiffusion and Diffmv leverage diffusion for EHR data augmentation, MedDiffusion is not well-suited to handle randomly missing views, as it generates different views independently. This reduces its dataset diversity.

Regarding task complexity, LOS Pred is particularly challenging, as it must correctly differentiate between closely related classes, often requiring more nuanced decision boundaries. Additionally, the fewer available entities per visit may exacerbate cold-start issues. Methods that rely on complete data, such as AdaCare and FlexCare, experience a notable performance degradation on this task. 
M3Care and PRISM demonstrate robust gains in both tasks, benefiting from knowledge completion brought about by the mapping.
Our method, Diffmv, outperforms state-of-the-art baselines with a minimum improvement of 2\%, primarily due to the generation of missing perspectives that enrich the data. Notably, cCGAN struggles to ensure consistency and stable generation, resulting in diminished performance when faced with the complexities of this task.

When considering the datasets, MIMIC-IV-Note presents the greatest challenges due to additional modality gaps between different views; for instance, notes are in text format while other EHR views are encoded as numerical codes. Algorithms designed for view concatenation, such as StageNet and HITANet, exhibit limited performance on this dataset. In contrast, GraphCare and PRISM demonstrate greater adaptability across multiple datasets, primarily due to their consideration of similar patients' representations. Further, view laziness constrains SMART and MedDiffusion's performance. As evidenced in  Figure~\ref{fig:case:inverse}, our approach effectively addresses this issue through inverse advantage reweighting, which balances the effectiveness of each view's utilization, resulting in improved overall performance.
\subsection{Model Analysis and Robust Testings}\label{sec:4.3}
Without loss of generality, we do robustness tests on MIMIC-III. 

\subsubsection{Ablation Study}\label{sec:4.3.1}
We conduct targeted ablation studies on the designed submodules while keeping the other modules intact like~\cite{jiang2023graphcare,xu2023seqcare}. As shown in Table~\ref{tab:aba}, removing any submodule leads to a performance decline, with Diffmv-NM experiencing a 2\% drop in F1-score (PHE). This arises from the absence of essential view transformation relationships, hindering the model's understanding of the missing view and causing over-reconstruction of the existing view. Removing crucial conditions in Diffmv-NIA and -NIT results in Jaccard decline (PHE) of 4\% and 2\%, respectively, highlighting the need to incorporate longitudinal disease evolution and relevant diagnostic considerations. Diffmv-NA exhibits slight accuracy degradation and a 3\% F1-score decline (LOS) due to potential bias from dominant views. Diffmv-NC removes the constraint of aligning the conditional sampling distribution with a normal distribution during denoising, which may lead to overfitting, akin to~\cite{ho2021classifier}. Diffmv-T outperforms the vanilla transformer in Table~\ref{tab:phe} and~\ref{tab:los} due to its ability to utilize contextual relationships for better imputations. However, it has limitations. Unlike diffusion, the transformer cannot effectively model gradual dependencies among views and may overlook the nuanced representation of uncertainty.
Overall, our findings demonstrate that each submodule plays an indispensable role in the final generation and prediction processes.
\begin{table}[!h]\small
\vspace{-1em}
\centering
\setlength{\abovecaptionskip}{-0.05cm}   
\caption{Ablation study. Diffmv-NM does not use an explicit binary mask and instead diffuses all representations simultaneously. Diffmv-NIA lacks intra-patient consistency, while Diffmv-NIT does not account for inter-patient consistency. Diffmv-NA eliminates the reweighting and employs concatenation to fuse all views. Diffmv-NC does not implement a classifier-free supervision. Diffmv-T trains a transformer to impute missing views without using diffusion.
} 
\label{tab:aba}
\resizebox{0.48\textwidth}{!}{
\begin{tabular}{c|c|cccccc||c} 
\hline
Algorithms                & Metric & -NM  & -NIA  & -NIT  & -NA & -NC & -T & Diffmv  \\ 
\hline
\multirow{2}{*}{PHE Pred}      & Jaccard  & 0.2716 & 0.2721 & 0.2773 & 0.2747    & 0.2808  & 0.2692 & \textbf{0.2831}       \\
                          & F1-score    & 0.4402 & 0.4468 & 0.4503 & 0.4275    & 0.4472  & 0.4315 & \textbf{0.4534}        \\ 
\hline\hline
\multirow{2}{*}{LOS Pred} & Accuracy & 0.4331 & 0.4399  & 0.4424 & 0.4400    & 0.4425 & 0.4359 & \textbf{0.4468}      \\
                          & F1-score  & 0.3854 & 0.3947  & 0.4008 & 0.3975    & 0.4054   & 0.3976 &\textbf{0.4100}      \\
                          
\hline
\end{tabular}}
\vspace{-1em}
\end{table}
\subsubsection{Warm-cold Examination}\label{sec:4.3.2}
\begin{figure}[!h] 
\vspace{-0.1cm}
\centering
\subfigure[PHE Pred (Jaccard)]{
\begin{minipage}[t]{0.48\linewidth}
\centering
\includegraphics[width=\linewidth,height=0.7\linewidth]{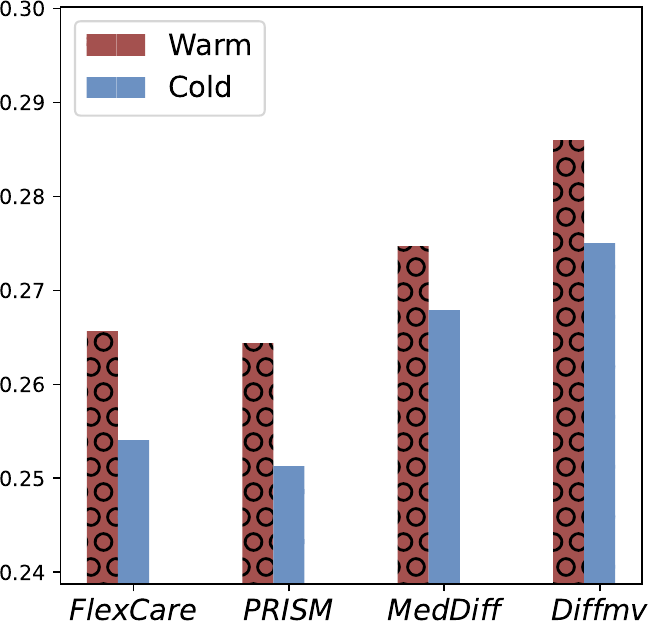}
\label{fig:warm:jac}
\end{minipage}%
}%
\subfigure[LOS Pred (Accuracy)]{
\begin{minipage}[t]{0.48\linewidth}
\centering
\includegraphics[width=\linewidth,height=0.7\linewidth]{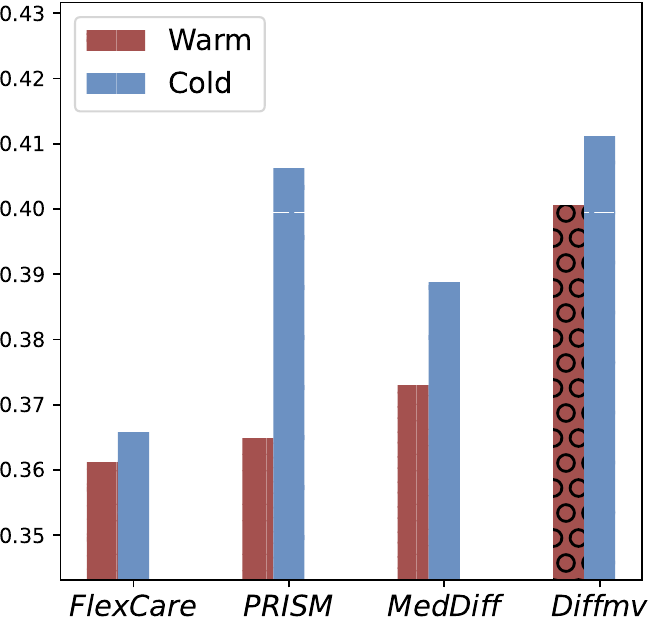}
\label{fig:warm:acc}
\end{minipage}%
}%
\centering
\setlength{\abovecaptionskip}{-0.05cm}   
\setlength{\belowcaptionskip}{-0.1cm}   
\caption{Warm-cold examination. To enhance the figure's appearance, MedDiffusion is abbreviated as MedDiff.} 
\label{fig:warm}
\vspace{-0.3cm}
\end{figure}

We further examine the performance enhancement for warm patients, defined as those with more than two visits, and cold patients, defined as those with only one visit. As depicted in Figure~\ref{fig:warm}, our analysis yields two key findings. First, we observe a decline in all algorithms' predictive performance for cold patients in PHE Pred. This decline can be attributed to the limited longitudinal data available for cold patients, which substantially reduces the richness of the information. However, counterintuitively, cold patients perform better on LOS Pred, a phenomenon we attribute to higher complications uncertainty and lower outcome dependency, as detailed in Appendix~\ref{app:anom}.
Second, we find that Diffmv demonstrates greater performance advantage in the warm group. This is because more EHR data enhances our understanding of patients' medical conditions, leading to more accurate imputations. This advantage is further amplified in both groups for PHE Pred, stemming from the longer sequences and richer entity interactions available. 

\subsubsection{Group Analysis of Missing Status}\label{sec:4.3.3}
\begin{figure}[!h] 
\vspace{-1em}
\centering
\subfigure[PHE Pred (Jaccard)]{
\begin{minipage}[t]{0.48\linewidth}
\centering
\includegraphics[width=\linewidth,height=0.7\linewidth]{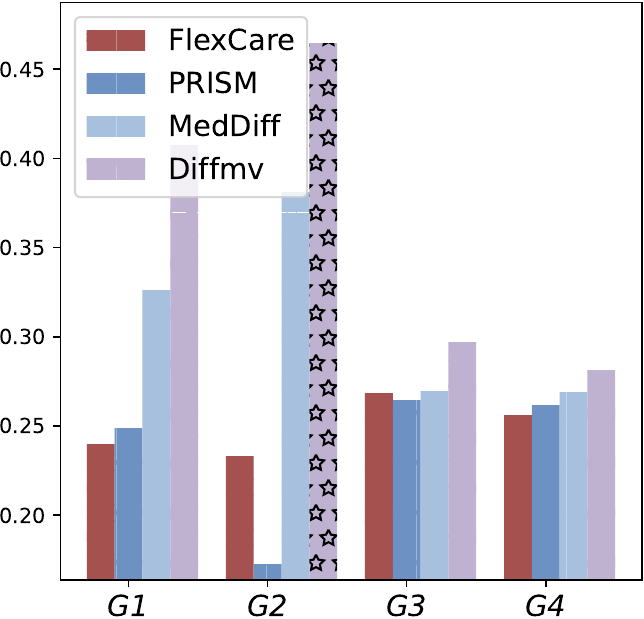}
\label{fig:gr:phe:jac}
\end{minipage}%
}%
\subfigure[LOS Pred (F1-score)]{
\begin{minipage}[t]{0.48\linewidth}
\centering
\includegraphics[width=\linewidth,height=0.7\linewidth]{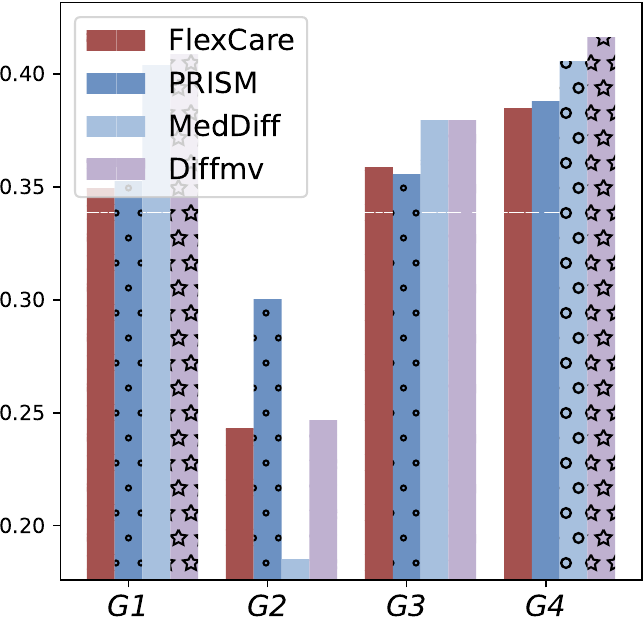}
\label{fig:gr:los:f1}
\end{minipage}%
}%
\centering
\setlength{\abovecaptionskip}{-0.05cm}   
\setlength{\belowcaptionskip}{-0.1cm}   
\caption{Group Analysis. G1-G4 represent patients with $(\frac{1}{2}, \frac{2}{3}]$,  $(\frac{1}{3}, \frac{1}{2}]$, $(\frac{1}{6}, \frac{1}{3}]$,$[0, \frac{1}{6}]$ missing view ratio. Please note that $\frac{2}{3}$ represents the maximum limit for missing ratio, as at least one view is retained for each visit.}
\label{fig:gr}
\vspace{-0.1cm}
\end{figure}

We further segment the patient group to examine the effectiveness of imputations for various missing statuses. G1 is the group with the highest missing level. As shown in Figure~\ref{fig:gr}, our findings indicate that imputation significantly aids in understanding G1-G4 patients' physical condition, aligning with our initial expectations. 
A detailed examination reveals that, for FlexCare and PRISM—lacking generative imputation—model performance improves as the missing level decreases, due to the richer information from more complete datasets. In contrast, MedDiffusion and Diffmv show significant gains in G1 and G2, attributed to the information gain from their imputation methods.
In G4, improvements from generative methods are modest. This is likely due to the higher information completeness, leading models to rely more on existing data, and the richness of their EHRs resulting in more defined outputs with limited additional diversity.

\subsubsection{Case Study}\label{sec:4.3.5}
We conduct case studies to showcase the quality of our generations and highlight our approach's superiority.

\begin{figure}[!h] 
\vspace{-1em}
\centering
\subfigure[Frequency Distribution]{
\begin{minipage}[t]{0.48\linewidth}
\centering
\includegraphics[width=\linewidth,height=0.8\linewidth]{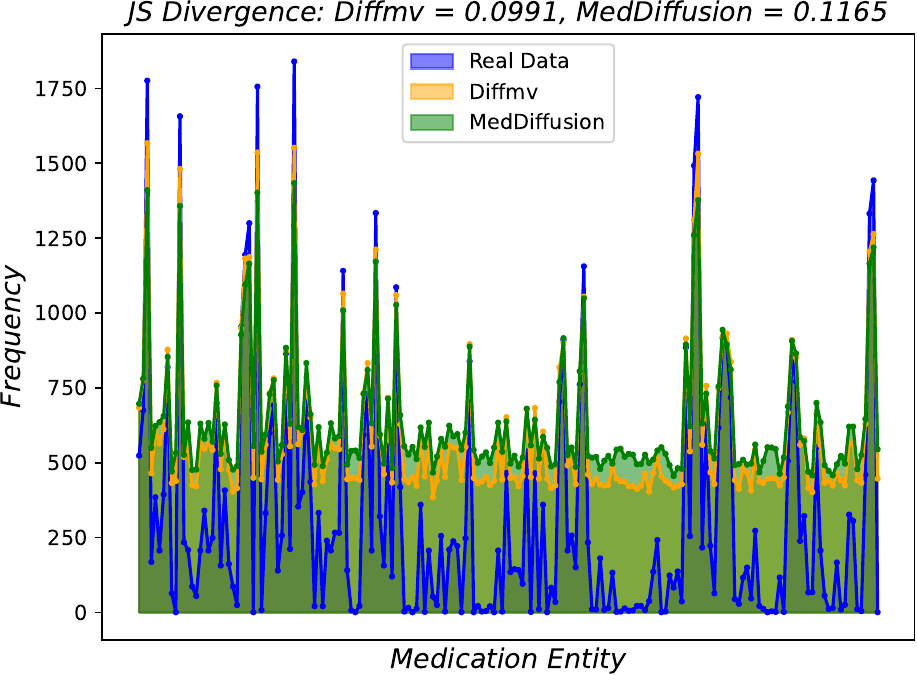}
\label{fig:case:jsd}
\end{minipage}%
}%
\subfigure[Correlation Heat Map]{
\begin{minipage}[t]{0.48\linewidth}
\centering
\includegraphics[width=\linewidth,height=0.8\linewidth]{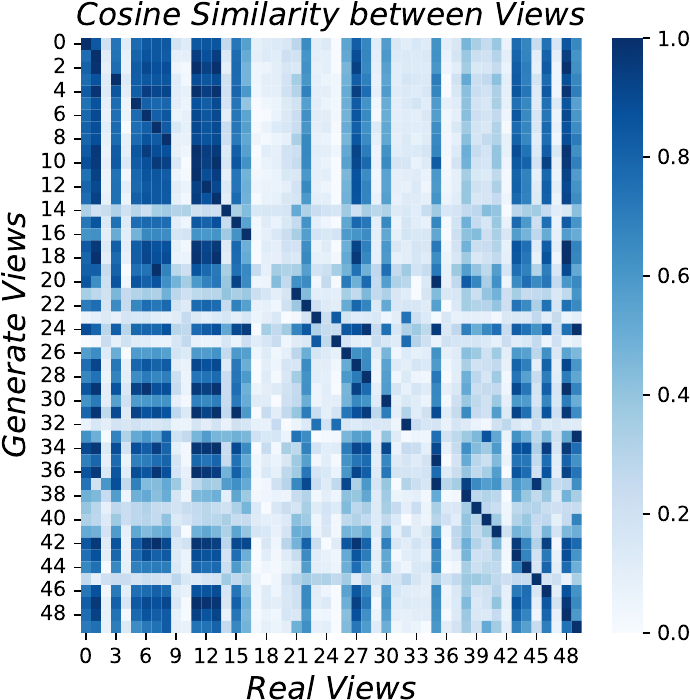}
\label{fig:case:heat}
\end{minipage}%
}%
\centering
\setlength{\abovecaptionskip}{-0.05cm}   
\setlength{\belowcaptionskip}{-0.1cm}   
\caption{Margin distribution generation. For Figure~\ref{fig:case:jsd}, we present the frequency distributions of generated and real entities from the medication view. For Figure~\ref{fig:case:heat}, We randomly select 50 missing views.}
\label{fig:case}
\vspace{-0.1cm}
\end{figure}

\noindent\textbf{Margin Distribution Match.}
To assess the alignment of our generated sample distribution with real samples, we conduct two experiments.
Specifically, we select complete visits from the test set and randomly mask a subset of the views. We then assess the distributional differences between the generated views and the masked views. As illustrated in Figure~\ref{fig:case:jsd}, our analysis reveals that the frequency distribution of the generated representations closely aligns with the true distribution (lower Jensen-Shannon Divergence~\cite{fuglede2004jensen}), outperforming MedDiffusion in this regard.
Furthermore, Figure~\ref{fig:case:heat} presents a heatmap illustrating the correlation between the different generated views and their corresponding real views. Notably, a higher similarity is observed along the diagonal of the heatmap, indicating that our model effectively produces coherent and distinguishable representations. This correlation not only reinforces the effectiveness of our method in synthesizing views that are both realistic and meaningful but also highlights its ability to capture the underlying data structure, contributing to enhanced interpretability in clinical applications. Additional visualizations are in Appendix~\ref{app:case}.

\noindent\textbf{View Laziness.}
"Lazy" is relative; if one view's contribution is too large, other views become lazy. To validate the impact of view laziness, we have increased the dominant weight of a specific view on MIMIC-III. The worse results validate our motivation—specifically-the risk of over-reliance on one view.
As depicted in Table~\ref{tab:view:la}, reducing the dominant view's weight from 10 to 2 effectively improve the performance, as it alters the contribution ratio among the various views. In our algorithm, this can be represented by assigning the dominant view a smaller rebalancing weight or increasing the rebalancing weight of lazy views. This experiment validates the rationale behind our motivation.
\begin{table}
\centering
\setlength{\abovecaptionskip}{-0.05cm}   
\caption{View Laziness. We do not use reweighting; instead, we increase the weight for a specific view. For PHE Pred, we use Jaccard, and for LOS Pred, we display AUROC.}
\label{tab:view:la}
\begin{tabular}{c|c|c|ccc} 
\toprule
Dominant                    & Task & Diffmv                                  & 2                                    & 5                                    & 10                                    \\ 
\hline
\multirow{2}{*}{Diagnose}   & PHE  & \textcolor[rgb]{0.2,0.2,0.2}{0.2831} & \textcolor[rgb]{0.2,0.2,0.2}{0.2460} & \textcolor[rgb]{0.2,0.2,0.2}{0.2302} & \textcolor[rgb]{0.2,0.2,0.2}{0.2010}  \\
                            & LOS  & \textcolor[rgb]{0.2,0.2,0.2}{0.8468} & \textcolor[rgb]{0.2,0.2,0.2}{0.8255} & \textcolor[rgb]{0.2,0.2,0.2}{0.8074} & \textcolor[rgb]{0.2,0.2,0.2}{0.7961}  \\ 
\hline\hline
\multirow{2}{*}{Procedure}  & PHE  & \textcolor[rgb]{0.2,0.2,0.2}{0.2831} & 0.2456                               & \textcolor[rgb]{0.2,0.2,0.2}{0.2180} & \textcolor[rgb]{0.2,0.2,0.2}{0.2067}  \\
                            & LOS  & \textcolor[rgb]{0.2,0.2,0.2}{0.8468} & \textcolor[rgb]{0.2,0.2,0.2}{0.8242} & \textcolor[rgb]{0.2,0.2,0.2}{0.8075} & \textcolor[rgb]{0.2,0.2,0.2}{0.7983}  \\ 
\hline\hline
\multirow{2}{*}{Medication} & PHE  & \textcolor[rgb]{0.2,0.2,0.2}{0.2831} & \textcolor[rgb]{0.2,0.2,0.2}{0.2479} & \textcolor[rgb]{0.2,0.2,0.2}{0.2252} & \textcolor[rgb]{0.2,0.2,0.2}{0.1826}  \\
                            & LOS  & \textcolor[rgb]{0.2,0.2,0.2}{0.8468} & \textcolor[rgb]{0.2,0.2,0.2}{0.8152} & \textcolor[rgb]{0.2,0.2,0.2}{0.8084} & \textcolor[rgb]{0.2,0.2,0.2}{0.7943}  \\
\bottomrule
\end{tabular}
\vspace{-1em}
\end{table}

\begin{figure}[!h] 
\centering
\includegraphics[width=\linewidth,height=0.21\linewidth]{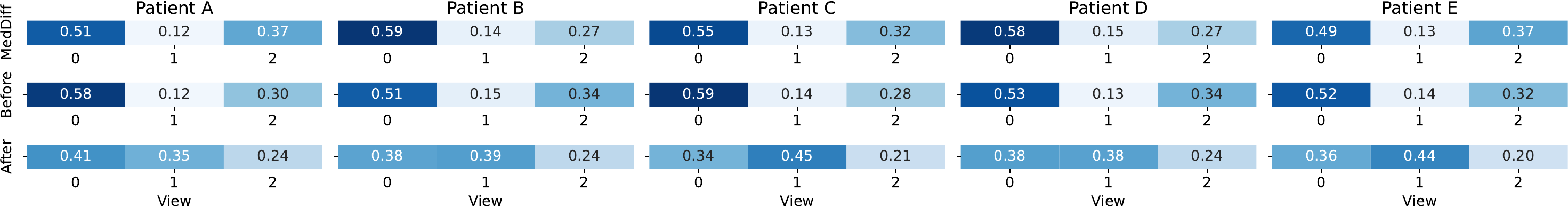}
\setlength{\abovecaptionskip}{-0.05cm}   
\setlength{\belowcaptionskip}{-0.1cm}   
\caption{Inverse Advantage Reweighting. The numbers represent the normalized gradients, while views 0, 1, and 2 refer to the diagnosis, procedure, and medication, respectively.}
\label{fig:case:inverse}
\vspace{-0.1cm}
\end{figure}

\noindent\textbf{Inverse Weight Rebalance.}
To intuitively illustrate the effect of the inverse advantage reweighting mechanism, we randomly select five patients to analyze their gradient contributions, a metric that assesses the impact of individual views on the final optimization~\cite{peng2022balanced}. As shown in Figure~\ref{fig:case:inverse}, the results indicate that after reweighting, the model achieves a more balanced gradient contribution among the views. This differs from MedDiffusion, which still exhibits an uneven distribution.
Our adjustment enhances the model's capacity to leverage diverse information sources, preventing any single view from dominating the final optimization. Consequently, this balance improves the overall performance and robustness of Diffmv.
\section{Conclusion}\label{sec:con}
In this study, we propose a novel generative framework to address the critical challenges of random missing views and view laziness. Our approach integrates multi-view EHRs within a unified diffusion denoising pipeline, employing a binary mask matrix to guide generation for diverse missing statuses. Furthermore, the innovative incorporation of intra- and inter-conditions ensures the consistency of patient health dynamics. We additionally introduce an inverse advantage reweighting strategy to promote the balanced utilization of various views. Extensive experiments demonstrate the robustness and superiority of our framework, highlighting its potential to improve patient care outcomes. In the future, we will validate our approach across a broader range of tasks.



\flushcolsend

\bibliographystyle{ACM-Reference-Format}
\vfill\eject
\balance
\bibliography{main}
\newpage
\clearpage
\appendix

\section{Missing Paradigm Difference}\label{app:diff}
\begin{figure}[!h] 
\vspace{-1em}
\centering
\begin{minipage}[t]{\linewidth}
\centering
\includegraphics[width=\linewidth,height=0.4\linewidth]{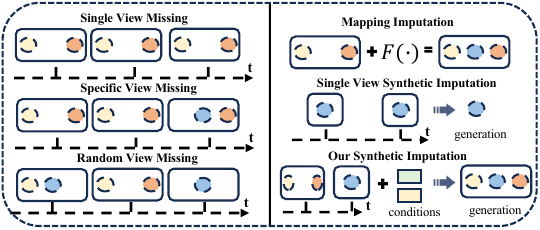}
\end{minipage}%
\centering
\setlength{\abovecaptionskip}{-0.05cm}   
\setlength{\belowcaptionskip}{-0.1cm}   
\caption{Paradigm difference. Note that this diagram does not describe more technical details, but only observes the essential difference in missing.}
\label{fig:diff}
\vspace{-2em}
\end{figure}

From Figure~\ref{fig:diff}, we observe that random missing data presents greater challenges and aligns more closely with real Electronic Health Record (EHR) data. Furthermore, our generative paradigm notably differs from mapping-based discriminative paradigms, used by M3Care and PRISM. Simultaneously, our synthesis paradigm employs a unified diffusion framework for data synthesis while also accounting for various conditions. This approach stands in stark contrast to MedDiffusion, which relies on single-view diffusion and needs multiple diffusion methods for random view missing.

\section{Theoretical Derivation}\label{app:theory}
In this section, we provide a detailed derivation of the formulas used in the forward and reverse processes. This aims to help readers quickly grasp the foundational concepts of diffusion and the key formula transformations presented in Section~\ref{sec:3.1}.

\noindent\textbf{Forward Diffusion Process.}
The forward process of diffusion aims to gradually add $T$-step noise to the data, introducing randomness that prepares it for the subsequent reverse denoising process. Following~\cite{ruan2023mm,peebles2023scalable,gong2022diffuseq}, we have $q(\mathbf{h}_t \mid \mathbf{h}_{t-1})=\mathcal{N}(\mathbf{h}_t ; \sqrt{1-\beta_t} \mathbf{h}_{t-1}, \beta_t \mathbf{I})$, with $t =1,2,\dots,T$ and $\{\beta_{t} \in (0,1)\}_{1}^{T}$. Let $\alpha_{t}=1-\beta_{t}$ and $\bar{\alpha}_{t}=\prod_{i=1}^t \alpha_i$. We use the Markov chain criterion to simplify the forward noise addition. Formally, 
\begin{equation}\label{eq:20}
\begin{aligned}
\mathbf{h}_t & =\sqrt{\alpha_t} \mathbf{h}_{t-1}+\sqrt{1-\alpha_t} \epsilon_{t-1}=\sqrt{\alpha_t \alpha_{t-1}} \mathbf{h}_{t-2}+\sqrt{1-\alpha_t \alpha_{t-1}} \bar{\epsilon}_{t-2} \\
& =\ldots=\sqrt{\bar{\alpha}_{t}} \mathbf{h}_0+\sqrt{1-\bar{\alpha}_t} \epsilon
\end{aligned},
\end{equation}
where $\epsilon$ refers to the Gaussian noises.
In the end, we could get,
\begin{equation}\label{eq:21}
q(\mathbf{h}_t \mid \mathbf{h}_0)=\mathcal{N}(\mathbf{h}_t ; \sqrt{\bar{\alpha}_t} \mathbf{h}_0,(1-\bar{\alpha}_t) \mathbf{I}).
\end{equation}
This allows us to eliminate cumbersome iterative steps and perform noise addition in a single step.

\noindent\textbf{Reverse Denoising Process.} 
The reverse process aims to reconstruct the original representation from the noisy data~\cite{li2022diffusion,peebles2023scalable,gong2022diffuseq}. Formally, 
\begin{equation}\label{eq:22}
\begin{aligned}
p_\theta(\mathbf{h}_{0: T})&=p(\mathbf{h}_T) \prod_{t=1}^T p_\theta(\mathbf{h}_{t-1} \mid \mathbf{h}_t), \\
p_\theta(\mathbf{h}_{t-1} \mid \mathbf{h}_t)&=\mathcal{N}(\mathbf{h}_{t-1} ; \mu_\theta(\mathbf{h}_t, t), \sigma_\theta(\mathbf{h}_t, t)),
\end{aligned}
\end{equation}
where $p_{\theta}$ is learned by our denoising network $g_{\theta}(\mathbf{h}_{t},t)$ and $\mu_{\theta}(\cdot)$ and $\sigma_{\theta}(\cdot)$ are the predicted parameterization of the mean and standard variation of $q(\mathbf{h}_t \mid \mathbf{h}_{t-1})$.
According to Bayes’ rule, we could also get,
\begin{equation}\label{eq:23}
q(\mathbf{h}_{t-1} \mid \mathbf{h}_t, \mathbf{h}_0)=q(\mathbf{h}_t \mid \mathbf{h}_{t-1}, \mathbf{h}_0) \frac{q(\mathbf{h}_{t-1} \mid \mathbf{h}_0)}{q(\mathbf{h}_t \mid \mathbf{h}_0)},
\end{equation}
Substituting Eq.~\ref{eq:20} into the equation, we can further derive, 
\begin{equation}\label{eq:24}
\begin{aligned}
    q(\mathbf{h}_{t-1} \mid \mathbf{h}_t, \mathbf{h}_0) &\sim \mathcal{N}(\mathcal{A}\mathbf{h}_t+\mathcal{B}\mathbf{h}_0, \sigma_{t}),\\
\mathcal{A}=\frac{\sqrt{a_t}(1-\bar{a}_{t-1})}{1-\bar{a}_t} , \mathcal{B}=& \frac{\sqrt{\bar{a}_{t-1}}(1-a_t)}{1-\bar{a}_t}, \sigma_{t}=(\frac{\sqrt{1-a_t} \sqrt{1-\bar{a}_{t-1}}}{\sqrt{1-\bar{a}_t}})^2,
\end{aligned}
\end{equation}
where all parameters of $\sigma_{t}$ are known; therefore, we only need to consider the mean.
In the reverse step, we want to maximize $\mathbb{E}_{q}\left[\log p_\theta(\mathbf{h}_0)\right]$ to approximate the true denoising distribution. This is equal to using the variational lower bound to optimize the negative log-likelihood $\mathbb{E}_{q}\left[-\log p_\theta(\mathbf{h}_0)\right] \leq \underset{\theta}{\mathcal{L}_{\text{vlb}}}$. Formally, similar to Kullback-Leibler divergence combinations in~\cite{sohl2015deep},
\begin{equation}\label{eq:25}
\begin{aligned}
\underset{\theta}{\mathcal{L}_{\text{vlb}}}=&\mathcal{L}_T+\mathcal{L}_{T-1}+\cdots+\mathcal{L}_0  \\&=\mathbb{E}_{q(\mathbf{h}_{1: T} \mid \mathbf{h}_0)}\left[\log \frac{q(\mathbf{h}_T \mid \mathbf{h}_0)}{p_\theta(\mathbf{h}_T)}+\sum_{t=2}^T \log \frac{q(\mathbf{h}_{t-1} \mid \mathbf{h}_0, \mathbf{h}_t)}{p_\theta(\mathbf{h}_{t-1} \mid \mathbf{h}_t)}\right. \\
& \left.+\log \frac{q_\phi(\mathbf{h}_0 \mid \mathbf{w}^{d \oplus p \oplus m})}{p_\theta(\mathbf{h}_0 \mid \mathbf{h}_1)}-\log p_\theta(\mathbf{w}^{d \oplus p \oplus m} \mid \mathbf{h}_0)\right],
\end{aligned}
\end{equation}
where $\phi$ is the embedding layer, specifically referring to Eq.~\ref{eq:1}-\ref{eq:3} in our scenario.
Please note that the first and second terms can be obtained using Eq.~\ref{eq:24}, while the 3-th and 4-th terms involve additional embedding and rounding processes respectively. This is necessary because medical codes are discrete, and it is essential to ensure that the embeddings obtained in continuous space correspond to the appropriate semantics~\cite{li2022diffusion,peebles2023scalable}.

\noindent\textbf{Eq.~\ref{eq:8} $\rightarrow$ Eq.~\ref{eq:9}:} We follow~\cite{ho2020denoising}, using Eq.~\ref{eq:24} to further simplify the calculation of Eq.~\ref{eq:25}. Formally, for $t \in [1,T]$,
\begin{equation}
\resizebox{1.0\hsize}{!}{$
\begin{aligned}
\underset{\theta}{\mathcal{L}_t} & =\mathbb{E}_{\mathbf{h}_0}\left[\log \frac{q(\mathbf{h}_t \mid \mathbf{h}_0, \mathbf{h}_{t+1})}{p_\theta(\mathbf{h}_t \mid \mathbf{h}_{t+1})}\right]=\mathbb{E}_{\mathbf{h}_0}\left[\frac{1}{\mathcal{F}}||\mu_t(\mathbf{h}_t, \mathbf{h}_0)-\mu_\theta(\mathbf{h}_t, t)||^2\right] \\
& =\mathbb{E}_{\mathbf{h}_0}\left[\frac{1}{\mathcal{F}}||\mathcal{A} \mathbf{h}_t+\mathcal{B} \mathbf{h}_0-(\mathcal{A} \mathbf{h}_t+\mathcal{B} g_\theta(\mathbf{h}_t, t))||^2\right]\\
& =\frac{\mathcal{B}}{\mathcal{F}} \mathbb{E}_{\mathbf{h}_0}\left[||\mathbf{h}_0-g_\theta(\mathbf{h}_t, t)||^2\right],
\end{aligned}$}
\end{equation}
where $\mathcal{F}=2{||{\sigma_\theta}||^{2}}$, is a loss independent constant. Then, the optimization of the training loss in Eq.~\ref{eq:25} is transformed into,
\begin{equation}
\resizebox{1.0\hsize}{!}{$
\begin{aligned}
& \underset{\theta}{\mathcal{L}_{\text{vlb}}} = \left[||\mu(\mathbf{h}_T)||^2 + \sum_{t=2}^T ||\mathbf{h}_0 - g_\theta(\mathbf{h}_t, t)||^2 
 + ||{\text{EMB}}(\mathbf{w}^{d \oplus p \oplus m}) - g_\theta(\mathbf{h}_1, 1)||^2 \right. \\
 &\left. - \log p_\theta(\mathbf{w}^{d \oplus p \oplus m} \mid \mathbf{h}_0)\right] \\
& \rightarrow  \min_\theta \left[ \sum_{t=2}^T ||\mathbf{h}_0 - g_\theta(\mathbf{h}_t, t)||^2 + ||{\text{EMB}}(\mathbf{w}^{d \oplus p \oplus m}) - g_\theta(\mathbf{h}_1, 1)||^2  \right.\\
& \left.- \log p_\theta(\mathbf{w}^{d \oplus p \oplus m} \mid \mathbf{h}_0)\right]
\end{aligned}
$},
\end{equation}
where $\text{EMB}(\cdot)$ represents the embedding process in Eq.\ref{eq:1}-\ref{eq:3} in our scenario. $\phi$ in Eq.~\ref{eq:25} serves as the embedding layer in the denoising network and can be omitted here.
$\mu(\mathbf{h}_T)$ is omitted in the last equation due to its constant value.

\section{Mathmatical Notations}\label{app:math}
Annotations for essential symbols are shown in Table~\ref{tab:math}.
\begin{table}\small
\centering
\setlength{\abovecaptionskip}{-0.05cm}   
\setlength{\belowcaptionskip}{-0.1cm}   
\caption{Mathematical Notations.}
\label{tab:math}
\resizebox{0.48\textwidth}{!}{
\begin{tabular}{c|l} 
\toprule
\textbf{Notations}                                                                                                     & \textbf{Descriptions}                                              \\ 
\hline
$\mathcal{U}  $                                                                                            & any ehr dataset                                          \\
$\emptyset$                                                                                            & the missing view                                        \\
$\mathcal{D},~\mathcal{P},~\mathcal{M}$   & diagnosis, procedure, and medication set                                    \\
$\mathbf{d},~\mathbf{p},~\mathbf{m} $                                  & multi-hot code of diagnosis, procedure, and medication                    \\
$j$                                                                                            & $j$-th visit           \\
$\mathcal{T}$                                                                                            & length of visit          \\
\hline
$\mathbf{E}$                                                                                            & embedding function     \\
$\mathbf{e}$                                                                                            & code embedding     \\
$\mathbf{v}$                                                                                            & visit embedding     \\
$\mathbf{b}$                                                                                            & mask embedding     \\
$\mathbf{h}$                                                                                            & hidden state for diffusion     \\
$q(a|b), p(a|b)$                                                                                            & conditional probability
a | b; q for diffusion and p for denoise  \\
                                                                             
$\mathcal{N}(\cdot; \mu,\sigma)$                                                           & probability density function of a Gaussian with mean $\mu$ and variance $\sigma$.  \\ 

$\mathbb{E}$                                                            & mathematic expectation  \\ 
$T$                                                           & total diffusion steps  \\ 
$\mathbf{w}^{d}$                                                           & discrete medical code of view $d$  \\ 
$\mathbf{z}$                                                         & historical embeddings \\ 
${c}$, $\mathbf{c}$                                                           & prototype index, prototype embedding \\ 
$\mathbf{s}$                                                         & guidance scale \\ 

\hline
$\mathbf{C}$                                                                                      & contextual conditions                              \\
$\mathcal{E} $                                                                                  &  longitudinal encoder                                  \\
$\mathbf{u}^{d}$                                                                                      &   final embeddings of view $d$                            \\ 
$\mathbf{r}$,   $\mathbf{n}$                                                                                   &  relative advantage, inverse advantage weight                        \\ 
$\eta$                                                                                   &  balance weight                        \\ 
$\phi$,   $\theta$, $\Theta$                                                                                   &  parameters of embeddings, denoising network, longitudinal encoder          \\ 

\bottomrule
\end{tabular}}
\vspace{-1em}
\end{table}


\begin{table*}[!h]
\centering
\setlength{\abovecaptionskip}{-0.05cm}   
\setlength{\belowcaptionskip}{-0.1cm}   
\caption{Data Statistics across all datasets (PHE Pred || LOS Pred). Due to task-specific preprocessing variations, we present data statistics for all tasks. \# means the number of. avg. means the average of. "diag" "prod" and "med" are abbreviations for diagnosis, procedure, and medication, respectively.}
\label{tab:sta}
\resizebox{\textwidth}{!}{
\begin{tabular}{l|ccc||ccc} 
\toprule
\textbf{Items}                & \textbf{MIMIC-III}  & \textbf{eICU}     & \textbf{MIMIC-IV-Note}        & \textbf{\textbf{MIMIC-III}} & \textbf{\textbf{eICU}} & \textbf{\textbf{MIMIC-IV-Note}}  \\ 
\hline
\# of patients / \# of visits     &  7,500 / 12,412             &   2,893 / 3,281              &  62,712 / 177,585    &  46,518 / 58,934       & 164,454  / 193,602     &  190,258  /  454,124     \\
diag. / prod. / med. set size  &  4,256 / 1,332 / 193             &  1,086 / 401 / 1,409              &  17,814 / 6,992 / 202   &  6,985 / 2,033 / 198       &  1,691 / 463 / 1,412      & 26,144 / 12,789 / 202  \\
avg. \# of visits                      &  1.6549             &  1.1341           & 2.8318    &  1.2669             & 1.1772                &  2.3869            \\
avg. \# of diag history per visit               &  29.0559              &  12.3048     &  57.6069        & 17.1663               & 6.8964                & 50.5566          \\
avg.  \# of prod history per visit                & 7.3851              &  31.0171        &  8.0731           & 5.5495    &  26.1551                    & 7.2598           \\
avg. \# of med history per visit                 & 55.9485             &  14.4044      &  85.5739            & 34.4246    &   16.8468                      & 74.0592         \\
\bottomrule
\end{tabular}}
\vspace{-1em}
\end{table*}
\section{Data Statistics}\label{app:data}
We conduct experiments on three widely used multi-view EHR datasets: MIMIC-III~\cite{johnson2016mimic}, eICU~\cite{pollard2018eicu}, and MIMIC-IV-Note~\cite{johnson2023mimic}. MIMIC-III is a large, publicly available database with detailed patient information, while eICU focuses on diverse critical care data from multiple hospitals. MIMIC-IV-Note adds unstructured clinical notes to EHR data, enriching the contextual information.
MIMIC-III and eICU provide three views: diagnosis, procedures, and medications, whereas MIMIC-IV-Note includes clinical notes as an additional view. For entities in MIMIC-III and eICU, they are encoded in common ICD or ATC codes, while clinical notes are in free text format. All views may be incomplete, posing challenges for view utilization. Following prior work~\cite{jiang2023graphcare,zhao2024enhancing}, we retain only patients with at least one visit for MIMIC-III and eICU, and at least two visits for MIMIC-IV-Note, with specific statistical information summarized in Table~\ref{tab:sta}. Please note that each visit must include at least one view.

\section{Further Analysis}\label{app:fur}
These experiments provide insights into our scalability, interpretability, and parameter tuning.

\subsection{Plug-in Application}\label{sec:4.3.4}
We perform plug-in applications to examine our extensibility on the sequential encoder and pre-trained language model (PLM).

\begin{figure*}[!h] 
\centering
\subfigure[PHE Pred (Jaccard)]{
\begin{minipage}[t]{0.23\linewidth}
\centering
\includegraphics[width=\linewidth,height=0.7\linewidth]{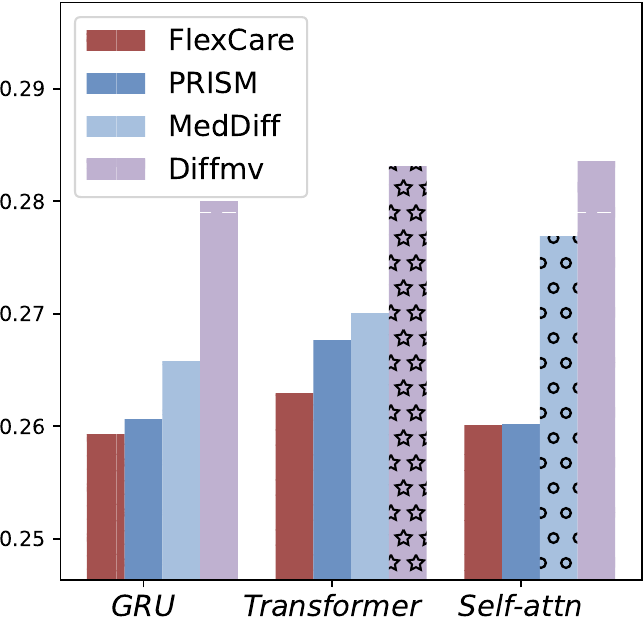}
\label{fig:plug:long:phe:jac}
\end{minipage}%
}%
\subfigure[PHE Pred (F1-score)]{
\begin{minipage}[t]{0.23\linewidth}
\centering
\includegraphics[width=\linewidth,height=0.7\linewidth]{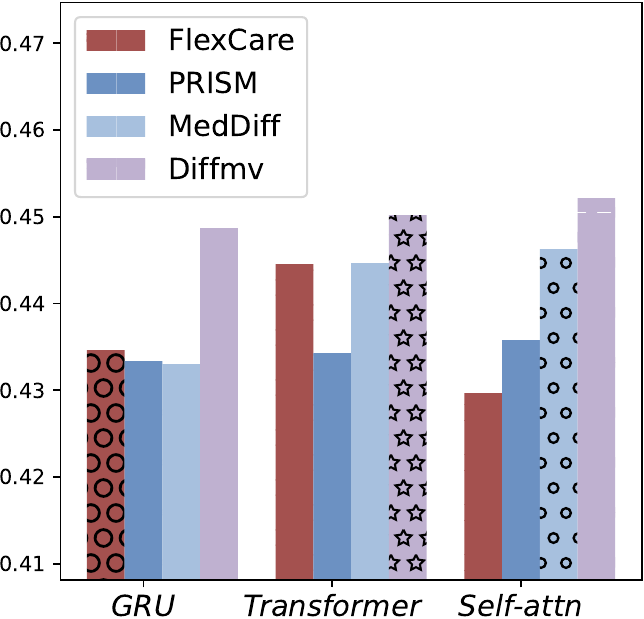}
\label{fig:plug:long:phe:f1}
\end{minipage}%
}%
\subfigure[LOS Pred (Accuracy)]{
\begin{minipage}[t]{0.23\linewidth}
\centering
\includegraphics[width=\linewidth,height=0.7\linewidth]{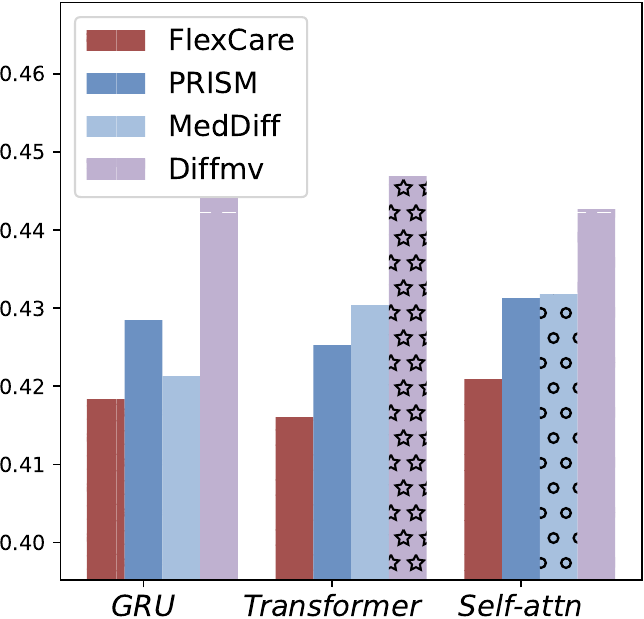}
\label{fig:plug:long:los:acc}
\end{minipage}%
}%
\subfigure[LOS Pred (F1-score)]{
\begin{minipage}[t]{0.23\linewidth}
\centering
\includegraphics[width=\linewidth,height=0.7\linewidth]{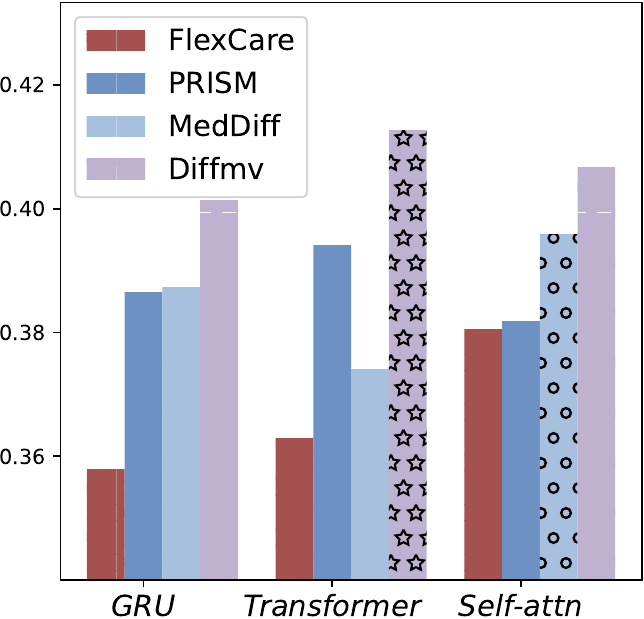}
\label{fig:plug:long:los:f1}
\end{minipage}%
}%
\centering
\setlength{\abovecaptionskip}{-0.15cm}   
\setlength{\belowcaptionskip}{-0.1cm}   
\caption{Plug-in (Diverse Longitudinal Encoder). We employ the popular GRU, Transformer, and Self-Attention for the analysis.}
\label{fig:plug:long}
\vspace{-0.1cm}
\end{figure*}

\noindent\textbf{Diverse Longitudinal Encoder.} 
In Section~\ref{sec:3.3}, we select the Transformer as the longitudinal representation encoder due to its versatility. We also evaluate other popular encoders: GRU~\cite{dey2017gate} and Self-Attention~\cite{usama2020self}. As illustrated in Figure~\ref{fig:plug:long}, FlexCare exhibits significant performance fluctuations, primarily due to its purely longitudinal paradigm. Weak encoders hinder effective temporal coordination capture, which critically affects performance. For example, Self-Attention does not explicitly encode positional information, which may result in the loss of such information.
In contrast, both MedDiffusion and our Diffmv effectively perceive the interaction patterns between EHRs during imputation, resulting in less performance degradation when using GRU and Self-Attention variants. 

\begin{figure*}[!h] 
\centering
\subfigure[PHE Pred (Jaccard)]{
\begin{minipage}[t]{0.23\linewidth}
\centering
\includegraphics[width=\linewidth,height=0.7\linewidth]{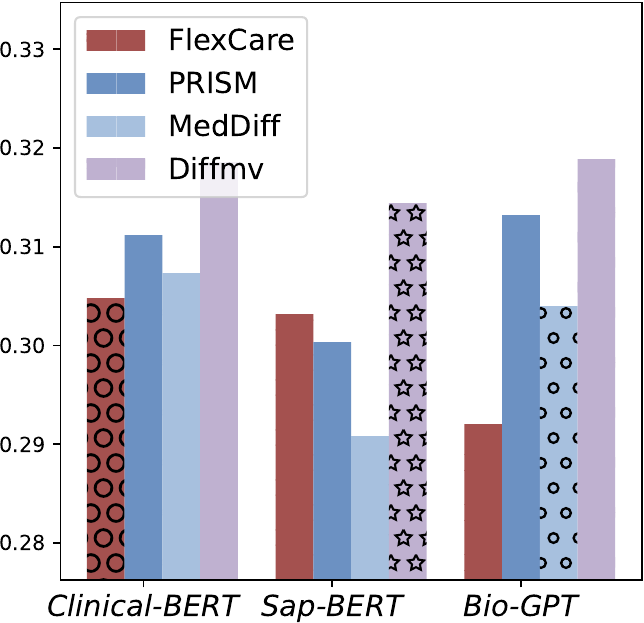}
\label{fig:plug:plm:phe:jac}
\end{minipage}%
}%
\subfigure[PHE Pred (F1-score)]{
\begin{minipage}[t]{0.23\linewidth}
\centering
\includegraphics[width=\linewidth,height=0.7\linewidth]{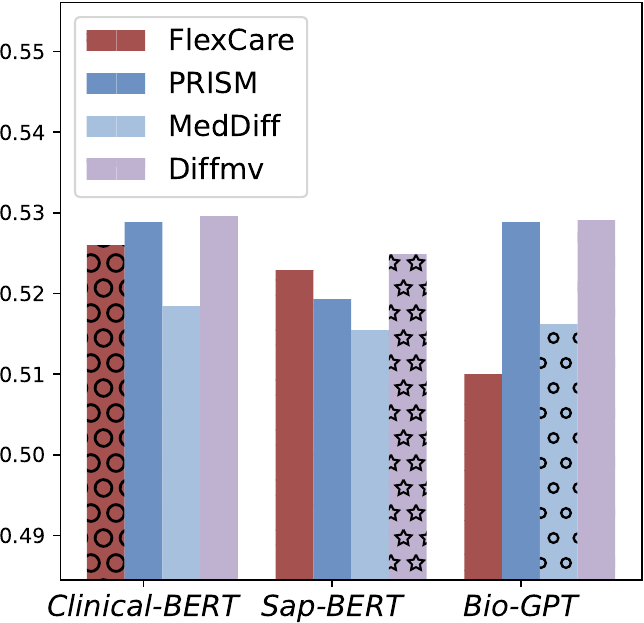}
\label{fig:plug:plm:phe:f1}
\end{minipage}%
}%
\subfigure[LOS Pred (Accuracy)]{
\begin{minipage}[t]{0.23\linewidth}
\centering
\includegraphics[width=\linewidth,height=0.7\linewidth]{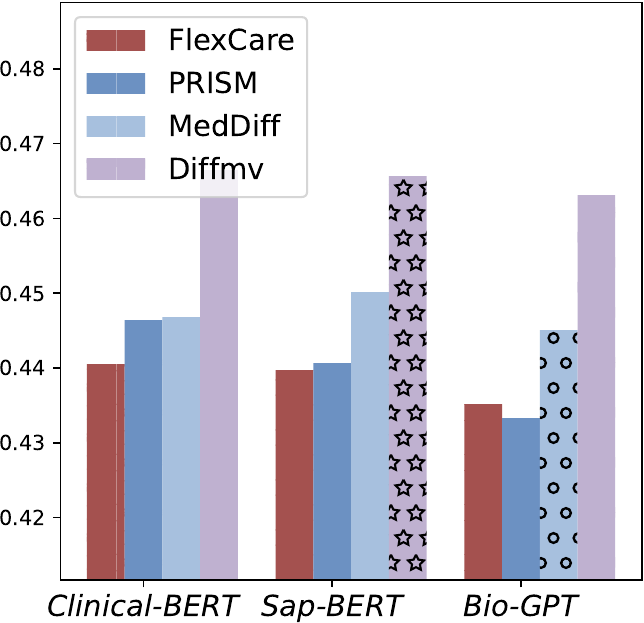}
\label{fig:plug:plm:los:acc}
\end{minipage}%
}%
\subfigure[LOS Pred (F1-score)]{
\begin{minipage}[t]{0.23\linewidth}
\centering
\includegraphics[width=\linewidth,height=0.7\linewidth]{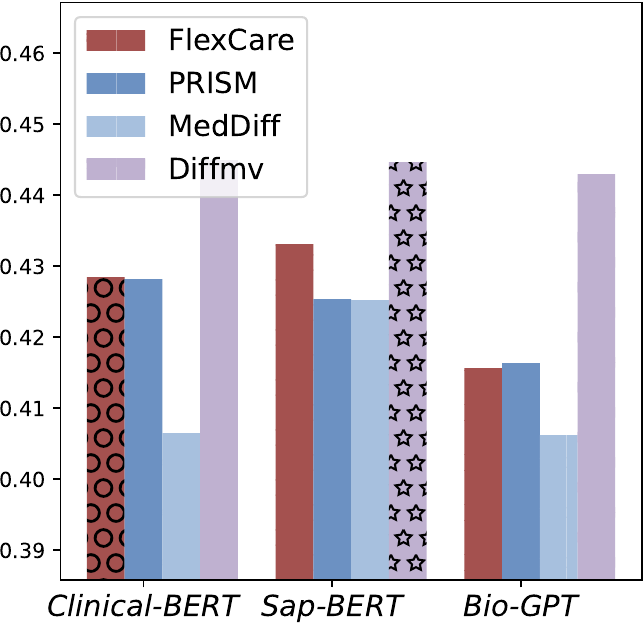}
\label{fig:plug:plm:los:f1}
\end{minipage}%
}%
\centering
\setlength{\abovecaptionskip}{-0.15cm}   
\setlength{\belowcaptionskip}{-0.1cm}   
\caption{Plug-in Application (Diverse PLM). We utilize the widely used Clinical-BERT, Sap-BERT, and Bio-GPT for the analysis.}
\label{fig:plug:plm}
\vspace{-0.1cm}
\end{figure*}

\noindent\textbf{Diverse PLM.}
In MIMIC-IV-Note, we utilize Sap-BERT as the pre-trained language model (PLM) for clinical notes, which is also pluggable. Additionally, we select Bio-BERT~\cite{lee2020biobert} and Clinical-BERT~\cite{wang2023optimized} for further experiments. As shown in Figure~\ref{fig:plug:plm}, our findings indicate that a more advanced PLM enhances the repair of missing views, with clear evidence of the superior performance of Clinical-BERT variants. This improvement may stem from the larger parameter sizes and richer corpora associated with this pre-train model, which create a more effective embedding space to capture the similarities among different EHR entities. Consequently, this opens up possibilities for progress in more complex multimodal scenarios.

In summary, these two experiments highlight our flexibility.

\subsection{Anomalies in Warm-cold Experiments.}\label{app:anom}
\begin{figure}[!h] 
\centering
\subfigure[Complications (LOS Pred)]{
\begin{minipage}[t]{0.45\linewidth}
\centering
\includegraphics[width=\linewidth,height=0.75\linewidth]{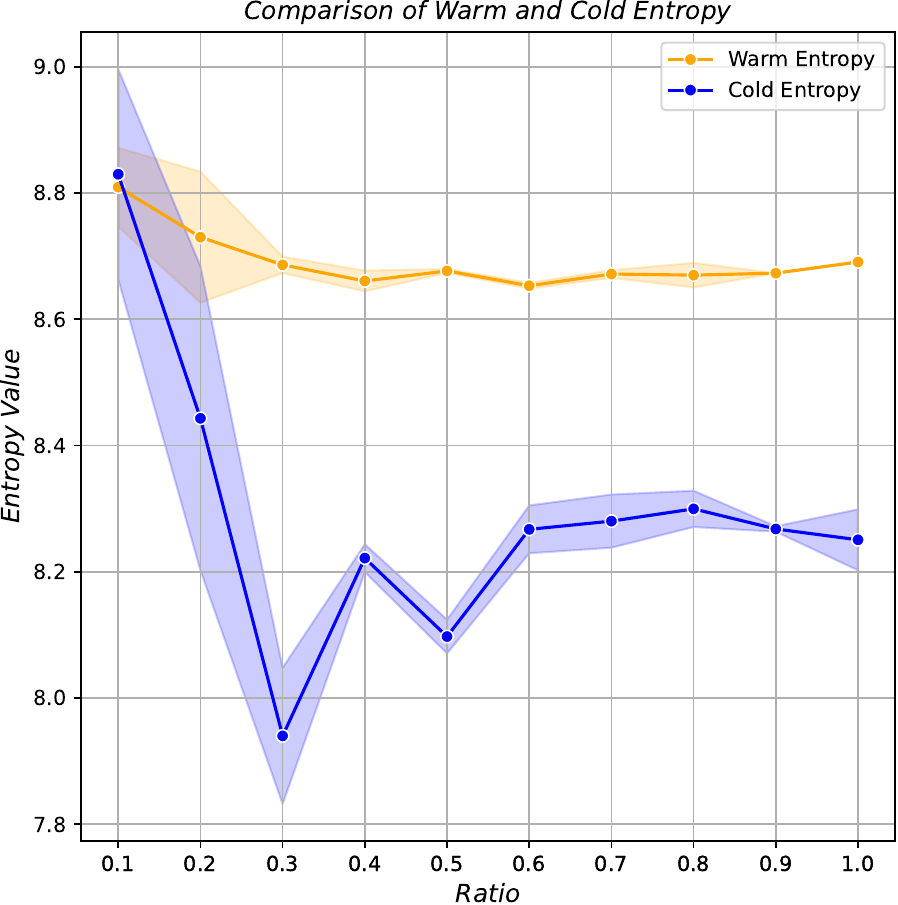}
\label{fig:case:com:los}
\end{minipage}%
}%
\subfigure[Outcome (LOS Pred)]{
\begin{minipage}[t]{0.45\linewidth}
\centering
\includegraphics[width=\linewidth,height=0.75\linewidth]{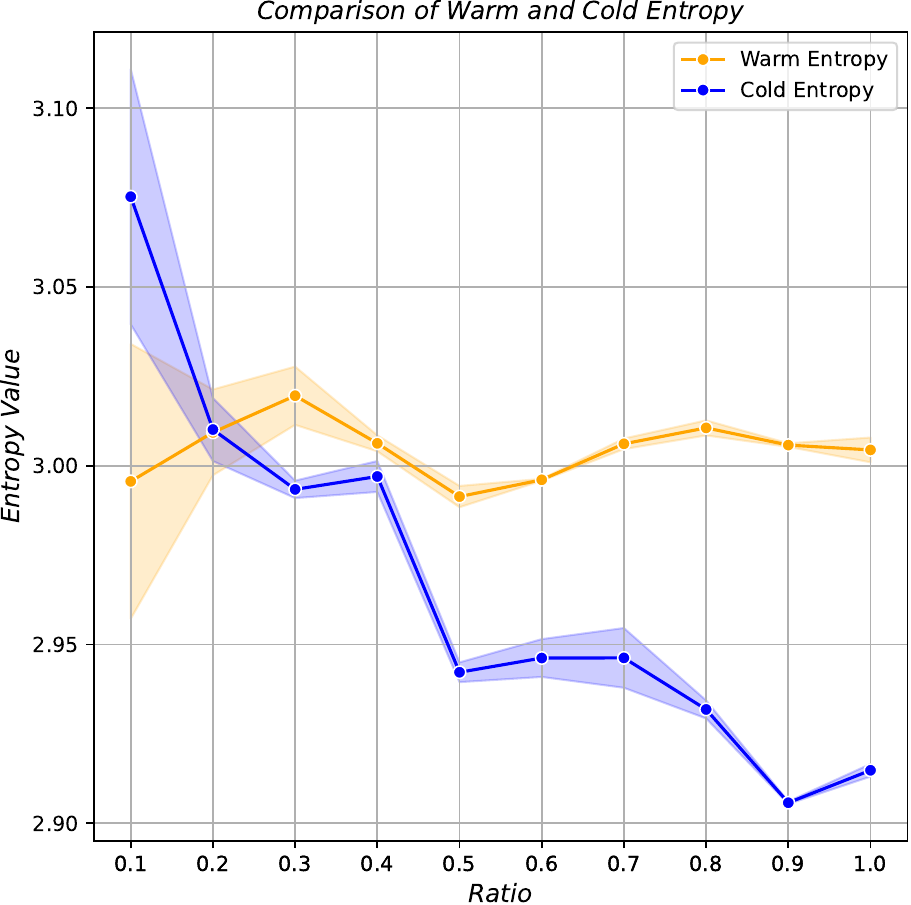}
\label{fig:case:out:los}
\end{minipage}%
}%
\centering
    \setlength{\abovecaptionskip}{-0.01cm} 
\caption{Figure~\ref{fig:case:com:los} depicts the entropy of complications corresponding to the warm / cold group in the training set. Figure~\ref{fig:case:out:los} showcases the entropy of labels corresponding to two groups in the training set. The horizontal axis represents the top proportion, where 0.1 indicates the top 10\% of diseases occurring in the group. The vertical axis denotes the entropy.
}
\label{fig:case:re}
\vspace{-1em}
\end{figure}
In Section~\ref{sec:4.3.2}, we observe that "warm" patients exhibit lower predictive accuracy in LOS Pred, which we attribute to significant differences in disease types between the groups, as illustrated in Figure~\ref{fig:case:re}. Specifically, we identify complications uncertainty and outcome dependency as potential reasons. Complications uncertainty is analyzed through complication entropy, where we sort diseases by occurrence frequency and record co-occurring complications. For instance, a disease in the top 10\% of the warm group with three complications occurring with counts \{1, 2, 3\} has its entropy calculated as "$\text{Entropy} = -\left( \frac{1}{6} \log \left( \frac{1}{6} \right) + \frac{2}{6} \log \left( \frac{2}{6} \right) + \frac{3}{6} \log \left( \frac{3}{6} \right) \right)$". Our findings, illustrated in Figure~\ref{fig:case:com:los}, indicate that warm patients experience greater complication entropy, suggesting a more diverse range of complications that increases the model's assessment complexity. Another perspective is outcome dependency, which examines whether labels remain consistent for specific diseases. Using a similar calculation method, we focus on label counts; for example, a disease in the top 10\% of the warm group with three associated labels (e.g., length of stay for LOS Pred) and counts \{1, 2, 3\} would yield a corresponding entropy. Higher entropy indicates lower label dependency, suggesting more diverse outcomes. Figure~\ref{fig:case:out:los} shows that cold patients have lower entropy, reflecting more defined treatment plans with aligned length of stay and medication protocols, making predictions easier to generalize. In contrast, warm patients exhibit higher entropy, indicating significantly different treatment plans, which complicates the model’s predictive performance in this group.

\subsection{Hyper-parameter Testing.}\label{app:hyper}
We further discuss several key hyperparameters.
\begin{figure}[!h] 
\vspace{-1em}
\centering
\subfigure[Prototype Num $K$]{
\begin{minipage}[t]{0.32\linewidth}
\centering
\includegraphics[width=\linewidth,height=0.7\linewidth]{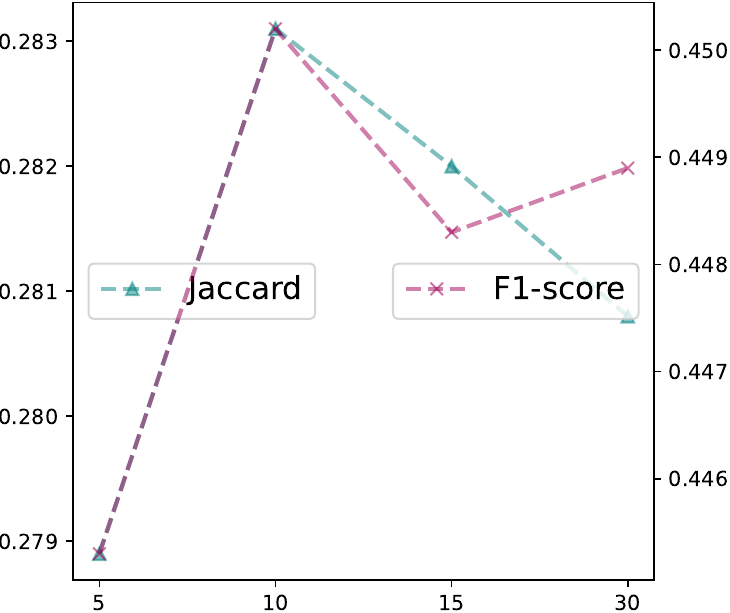}
\label{fig:hyper:pro}
\end{minipage}%
}%
\subfigure[Guidance Scale $s$]{
\begin{minipage}[t]{0.32\linewidth}
\centering
\includegraphics[width=\linewidth,height=0.7\linewidth]{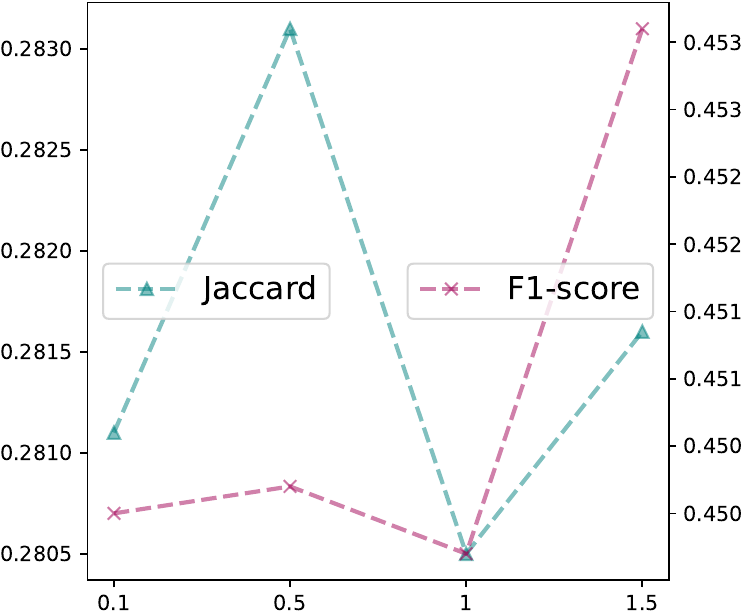}
\label{fig:hyper:gui}
\end{minipage}%
}%
\subfigure[Balance Weight $\eta$]{
\begin{minipage}[t]{0.32\linewidth}
\centering
\includegraphics[width=\linewidth,height=0.7\linewidth]{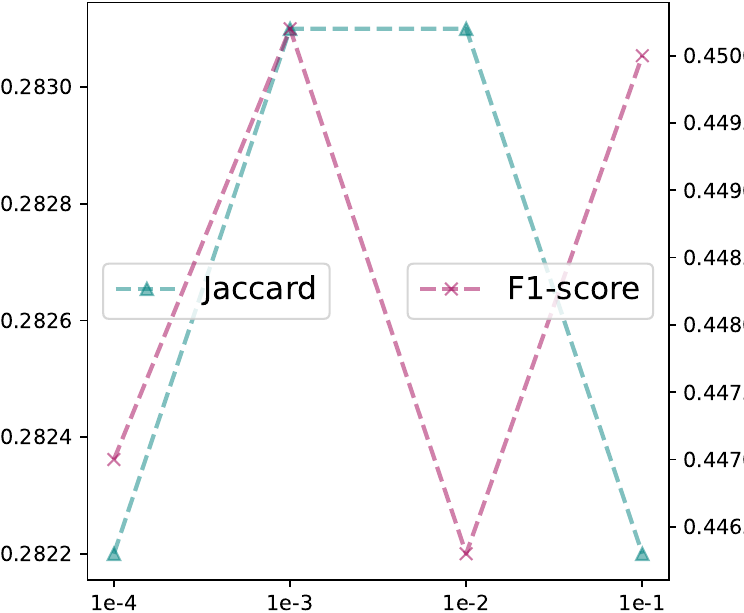}
\label{fig:hyper:wei}
\end{minipage}%
}%
\centering
\setlength{\abovecaptionskip}{-0.05cm}   
\setlength{\belowcaptionskip}{-0.1cm}   
\caption{(a) Performance under different prototype numbers (Num). (b) Performance under different guidance scales. (c) Performance under different balance weights. We show the all results on MIMIC-III (PHE Pred).}
\label{fig:hyper}
\vspace{-1em}
\end{figure} 

\noindent\textbf{Prototype Num $K$.}
The number of prototypes, as indicated in Eq.~\ref{eq:11}, refers to the coarse-grained clustering of overall visits. Intuitively, fewer prototypes lead to fewer clusters, indicating a higher degree of integration of similar patient concepts into the inter-patient consistency reference. This may result in incorrect guidance for patients with dissimilar symptoms.
Conversely, having more prototypes allows for the identification of more specific similar visits, thereby enabling more tailored diagnosis and treatment plans. However, this approach may yield a reference sample size that is too small and not representative.
As evidenced in Figure~\ref{fig:hyper:pro}, we select $K=10$.

\noindent \textbf{Guidance Scale $s$.}
As illustrated in the Figure~\ref{fig:hyper:gui}, our results indicate that when $s=0.5$.
Diffmv achieves optimal performance. The classifier-free guidance effectively minimizes the disparity between the conditional and unconditional generation results, thereby integrating the guidance of the conditional information without relying on an additional classifier. However, when the scale of the guidance is increased, the model may overlook the original diffusion loss and prioritize compatibility with contextual conditions, potentially leading to suboptimal outcomes.

\noindent \textbf{Balance Weight $\eta$.}
We evaluate the impact of the reweighting strategy in Figure~\ref{fig:hyper:wei}. When the weight is increased, the model emphasizes a more balanced utilization of each view; conversely, a lower weight allows the model to learn potential patterns more randomly. Interestingly, we find that greater balance does not necessarily yield better results. This may be because certain views are more valuable in specific contexts.
For instance, in cardiovascular diseases, the critical examination provided by the procedure is essential. In this context, clinical tests such as echocardiograms may carry more weight than medication history, as they directly influence treatment decisions. Overemphasizing balance may lead the model to incorporate unnecessary information from less relevant modalities, potentially diluting the effectiveness of crucial insights.

\section{Metric Definitions}\label{app:metric}

The evaluation metrics outlined in the paper are defined as follows,
\begin{equation}\label{eq:28}
    \text{Jaccard} = \frac{|\mathbf{y}_{\text{pred}} \cap \mathbf{y}_{\text{true}}|}{|\mathbf{y}_{\text{pred}} \cup \mathbf{y}_{\text{true}}|},
\end{equation}
where  
$\mathbf{y}_{\text{pred}}$ is the predicted set and $\mathbf{y}_{\text{true}}$ is the true set.

\begin{equation}\label{eq:29}
    \text{F1-score} = \frac{2 \times \text{precision} \times \text{recall}}{\text{precision} + \text{recall}},
\end{equation}
where precision and recall are defined as:
 $\text{precision} = \frac{\text{tp}}{\text{tp} + \text{fp}}$ and $\text{recall} = \frac{\text{tp}}{\text{tp} + \text{fn}}
$, where $\text{tp}$ is the number of true positives, 
$\text{fp}$ is the number of false positives, and 
$\text{fn}$ is the number of false negatives.

\begin{equation}\label{eq:30}
    \text{AUROC} = \int_{0}^{1} \text{tpr}(\text{fpr}) \, d(\text{fpr}),
\end{equation}
where tpr (True Positive Rate) and fpr (False Positive Rate) are calculated at different thresholds.

\begin{equation}\label{eq:31}
    \text{AUPRC} = \int_{0}^{1} \text{precision}(\text{recall}) \, d(\text{recall}),
\end{equation}
where precision and recall are evaluated at different thresholds.

\begin{equation}\label{eq:32}
\text{Accuracy} = \frac{\text{tp} + \text{tn}}{\text{tp} + \text{tn} + \text{fp} + \text{fn}},
\end{equation}

\begin{equation}\label{eq:33}
\text{Kappa} = \frac{p_o - p_e}{1 - p_e},
\end{equation}
where $p_o$ is the observed agreement (i.e., accuracy), and $p_e$ is the expected agreement by chance, calculated as:
\begin{equation}\label{eq:34}
p_e = \frac{(n_{\text{pos}} \cdot n_{\text{pred}} + n_{\text{neg}} \cdot n_{\text{true}})}{\tilde{N}^2},
\end{equation}
with $n_{\text{pos}}$ and $n_{\text{neg}}$ being the number of positive and negative instances, $n_{\text{pred}}$ and $n_{\text{true}}$ being the predicted counts of positive and negative instances, and $\tilde{N}$ being the total number of instances.

\section{Case Study}\label{app:case}
\noindent\textbf{Top-5 Generalization.}
We select the top five generated codes and examine the changes in their co-occurrence frequency rank in the training set after imputation. As demonstrated in Table~\ref{tab:gene}, the generated codes significantly alter the co-occurrence patterns observed in the original dataset. This alteration is intuitive, as visits with missing views can leverage auxiliary information from other views to infer the content. Consequently, the original partially sparse encoding experiences a marked enhancement in the number of co-occurrences. This change in distribution may help advance the model's capability in diagnosing and treating rare diseases or utilizing less common medications.
\begin{table}[!h]\small
\vspace{-1em}
\centering
\setlength{\abovecaptionskip}{-0.05cm}   
\caption{Top-5 Generation for medication view. $R(\cdot)$ refers to the rank of the code within the dataset. $D_{b}$ and $D_{a}$ denotes the medication entity before and after imputation.}
\label{tab:gene}
\resizebox{0.48\textwidth}{!}{
\begin{tabular}{cccc} 
\toprule
R($D_{b}$) & R($D_{a}$) & ATC-Code & Description  \\ 
\hline
78   & 10    & D07A      &   CORTICOSTEROIDS, OTHER COMBINATIONS     \\
85     & 17       & R07A         & OTHER RESPIRATORY SYSTEM PRODUCTS               \\
82     & 15       & A07F         &  ANTIDIARRHEAL MICROORGANISMS             \\
84   & 16       & D05B         &   ANTIPSORIATICS FOR SYSTEMIC USE           \\
83    & 21       & C05C         & CAPILLARY STABILIZING AGENTS             \\
\bottomrule
\end{tabular}}
\vspace{-0.1cm}
\end{table}

\begin{figure}[!h] 
\centering
\begin{minipage}[t]{0.8\linewidth}
\centering
\includegraphics[width=0.9\linewidth,height=0.7\linewidth]{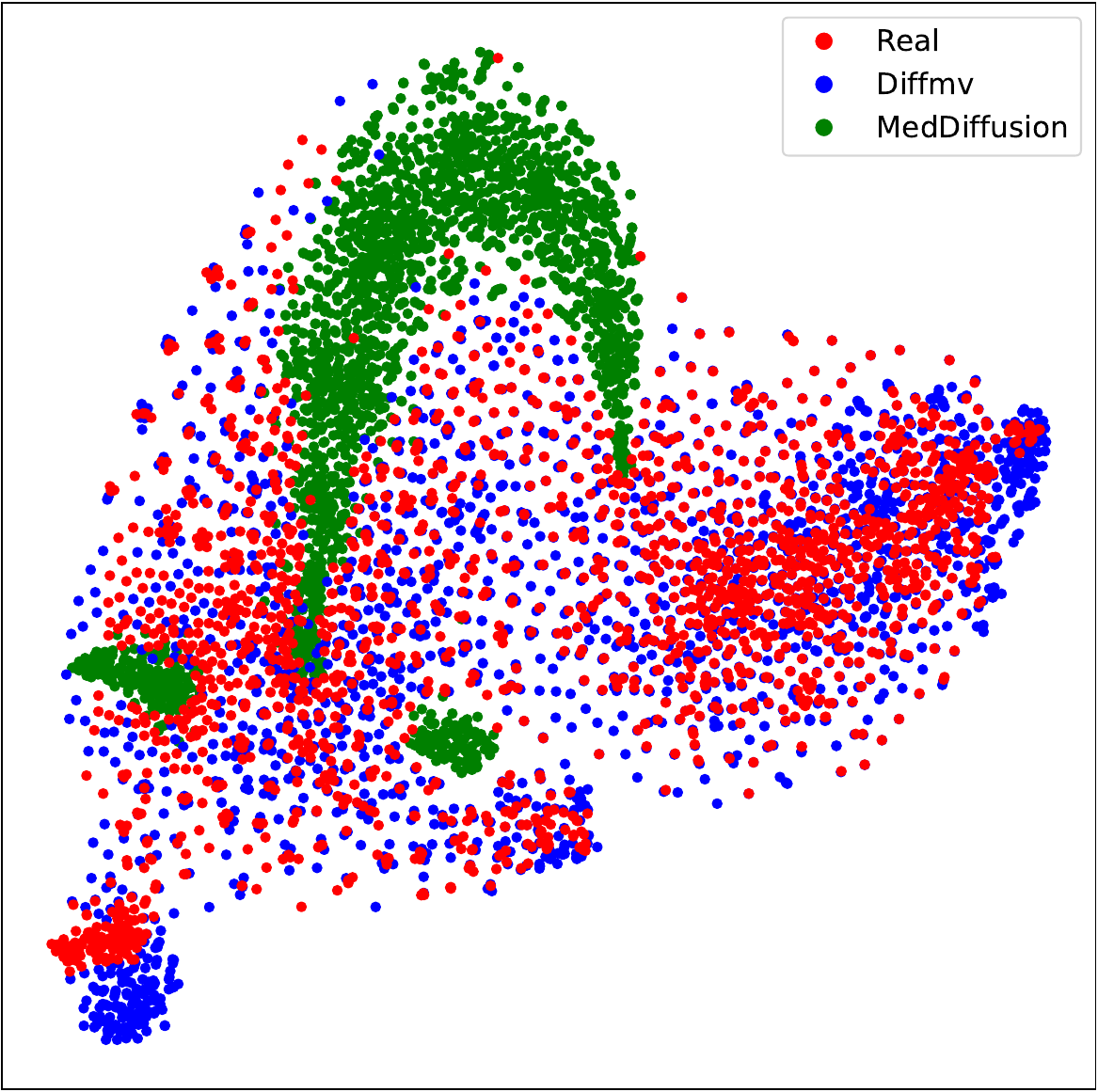}
\end{minipage}%
\centering
\setlength{\abovecaptionskip}{-0.05cm}   
\caption{T-SNE analysis. We visualize the difference between the representation distribution of the generated medication views and the real distribution.}
\label{fig:case:tsne}
\vspace{-1em}
\end{figure}

\noindent\textbf{T-SNE Analysis.}
As indirect evidence, we further extract visits with complete viewpoints from the test set and randomly mask them. We then compare the distribution of the generated sample representations with that of the masked true representations using T-SNE visualization~\cite{van2008visualizing}.
Our results in Figure~\ref{fig:case:tsne} indicate that Diffmv aligns more closely with the true distribution, highlighting the effectiveness of our approach. In contrast, MedDiffusion captures only a portion of the distribution, primarily due to its limitations in assimilating intra- and inter-patient consistency.
Overall, this visualization complements Section~\ref{sec:4.3.5} and offers a degree of interpretability.

\end{document}